\documentclass{article}

\usepackage{microtype}
\usepackage{graphicx}
\usepackage{subfigure}
\usepackage{booktabs} % for professional tables

\usepackage{hyperref}

\usepackage{amsmath}
\usepackage{amsfonts}
\usepackage{amssymb}
\usepackage{mathtools} 
\usepackage{bbm}
\usepackage{graphbox}

\usepackage{amsthm}
\theoremstyle{definition}
\newtheorem{definition}{Definition}[section]

\usepackage{wrapfig}
\usepackage{enumitem}
\usepackage{multirow}
\usepackage{caption}

\usepackage[accepted]{icml2022}

\icmltitlerunning{Robust Policy Learning over Multiple Uncertainty Sets}

\begin{document}

\twocolumn[
\icmltitle{Robust Policy Learning over Multiple Uncertainty Sets}



\icmlsetsymbol{equal}{*}

\begin{icmlauthorlist}
\icmlauthor{Annie Xie}{stanford}
\icmlauthor{Shagun Sodhani}{fair}
\icmlauthor{Chelsea Finn}{stanford}
\icmlauthor{Joelle Pineau}{fair}
\icmlauthor{Amy Zhang}{fair}
\end{icmlauthorlist}

\icmlaffiliation{stanford}{Stanford University}
\icmlaffiliation{fair}{Facebook AI Research}

\icmlcorrespondingauthor{Annie Xie}{anniexie@stanford.edu}

\icmlkeywords{Machine Learning, ICML}

\vskip 0.3in
]



\printAffiliationsAndNotice{}  % leave blank if no need to mention equal contribution

\begin{abstract}
Reinforcement learning (RL) agents need to be robust to variations in safety-critical environments. While system identification methods provide a way to infer the variation from online experience, they can fail in settings where fast identification is not possible. Another dominant approach is robust RL which produces a policy that can handle worst-case scenarios, but these methods are generally designed to achieve robustness to a single uncertainty set that must be specified at train time. Towards a more general solution, we formulate the multi-set robustness problem to learn a policy robust to different perturbation sets. We then design an algorithm that enjoys the benefits of both system identification and robust RL: it reduces uncertainty where possible given a few interactions, but can still act robustly with respect to the remaining uncertainty. On a diverse set of control tasks, our approach demonstrates improved worst-case performance on new environments compared to prior methods based on system identification and on robust RL alone.
\end{abstract}

\section{Introduction}
\label{sec:introduction}

\begin{figure}
\centering
\vspace{-0.3cm}
\includegraphics[width=\linewidth]{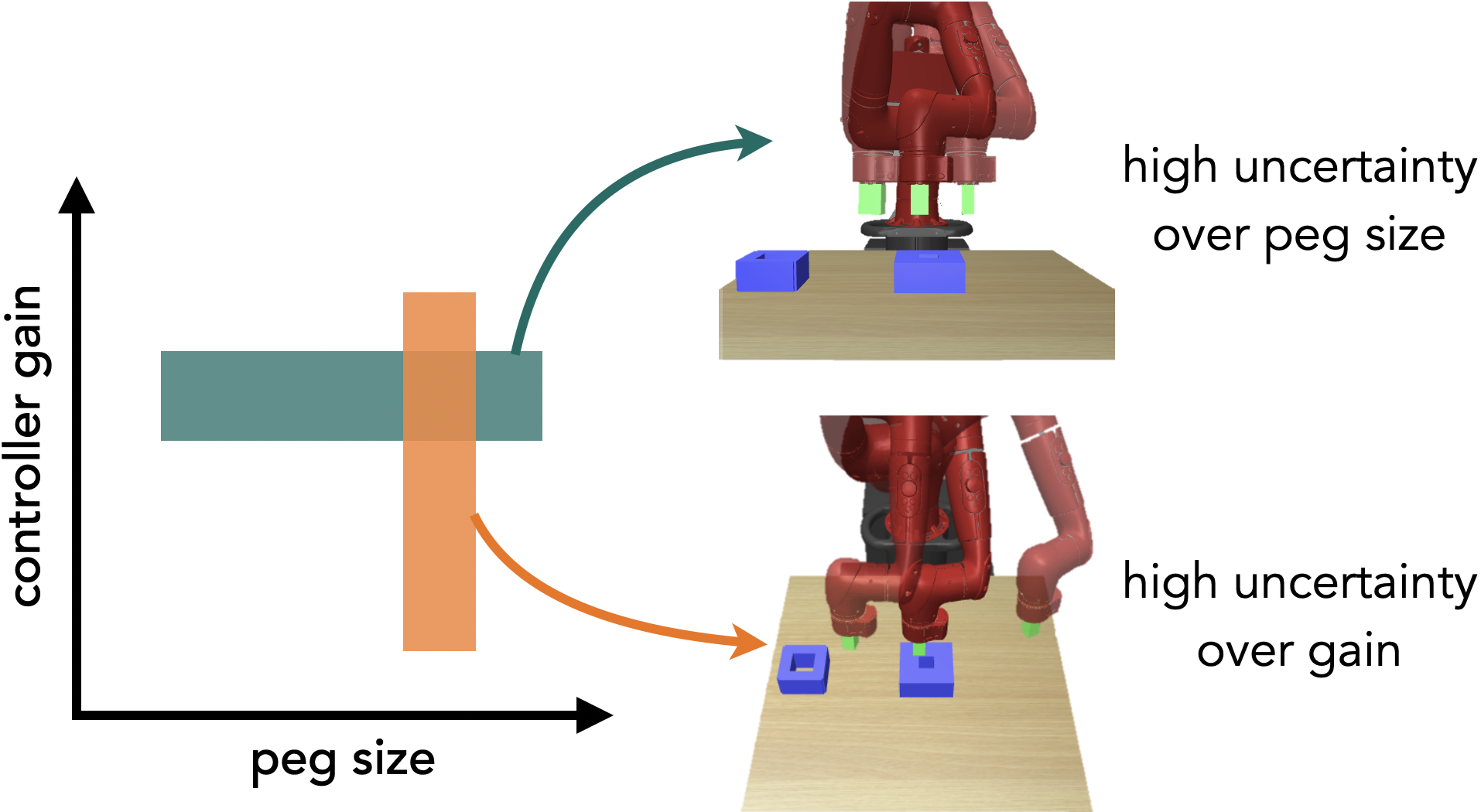}
\vspace{-0.7cm}
\caption{\small Two peg-insertion tasks. The first (green) presents high uncertainty in the peg size, while the second (orange) has high uncertainty in the controller gain.}
\label{fig:sawyer_peg}
\vspace{-0.5cm}
\end{figure}

Uncertainty is a prevalent challenge in most realistic reinforcement learning (RL) settings. Our work studies the uncertainty that arises when an agent is transferred to a new environment, after training on similar tasks related through a common set of underlying parameters, often referred to as the \emph{context}. In safety-critical settings, we often care about the agent's worst-case performance on the distribution of plausible environments. 

Robust RL is one of the primary approaches to this problem as it aims to learn a policy that performs well under worst-case perturbations to the context~\cite{rajeswaran2016epopt,pinto2017robust,mankowitz2019robust,tessler2019action,vinitsky2020robust,abraham2020model}. However, these solutions require a prior uncertainty set over the context for the \emph{test-time} environment to learn the robust policy for this set at training time. Building in this prior ahead of time can limit the flexibility of the resulting policy: a large uncertainty set produces an overly conservative policy that can potentially underperform in all environments, but a small uncertainty set can fail to represent the target environment~\citep{mozian2020learning}. 

\begin{figure}
    \centering
    \vspace{-0.1cm}
    \includegraphics[width=\linewidth]{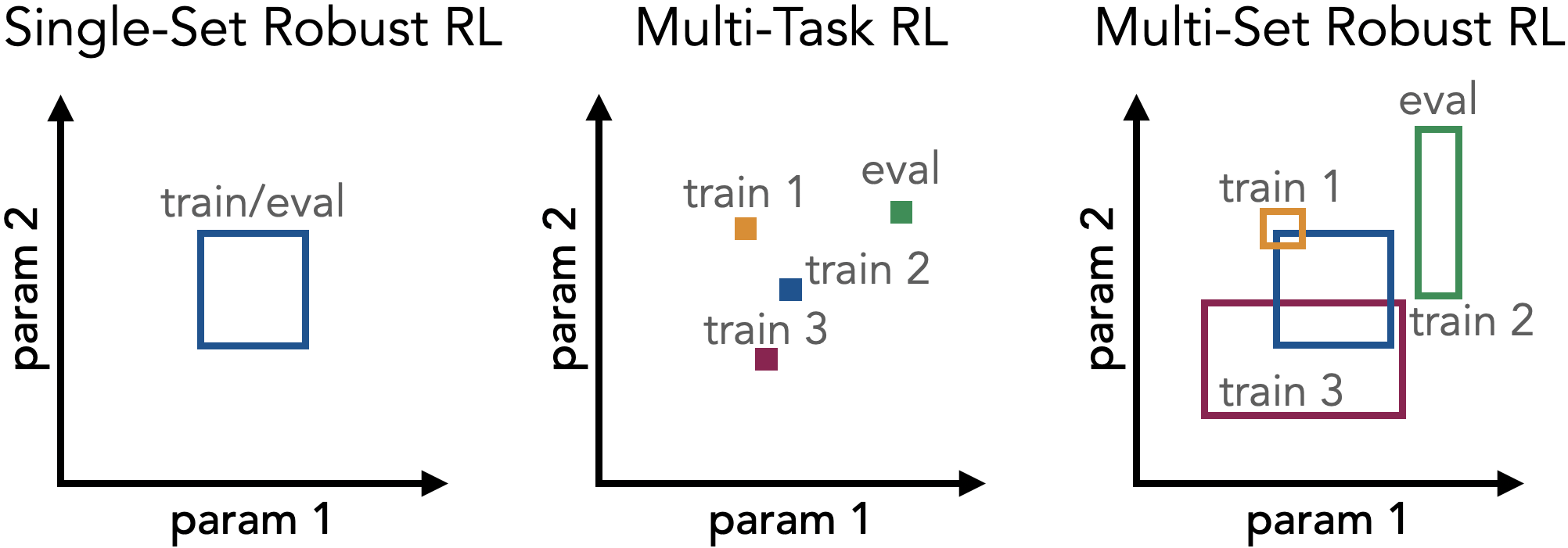}
    \vspace{-0.8cm}
    \caption{\small An illustration of the robust, multi-task, and multi-set robust RL setups. Robust RL learns a policy for a \emph{single uncertainty set}, while multi-task RL optimizes a policy to solve a \emph{collection of tasks}. Finally, multi-set robust RL aims to learn a policy that performs well with respect to a \emph{collection of uncertainty sets}.}
    \label{fig:problem}
    \vspace{-0.6cm}
\end{figure}

We, therefore, formulate and study the \emph{multi-set robustness} problem (illustrated in Fig.~\ref{fig:problem}) whose goal is to learn a policy with strong worst-case performance on new uncertainty sets. Since the optimal robust policy varies across different perturbation sets, we incorporate the uncertainty set as contextual information to the agent and learn a generalized set-conditioned policy. 

However, naively contextualizing existing robust methods with the uncertainty set can still be sub-optimal as these methods do not reduce uncertainty over the context. In particular, the parameters that make up the context can sometimes be quickly identified, given a history of interactions. 
For example, consider a robot inserting a peg in one of the boxes in Fig.~\ref{fig:sawyer_peg}. The box closer to the robot only fits smaller pegs, while the box to the left can accommodate all sizes. Hence, the optimal policy should select the closer box for smaller pegs and the faraway box for larger ones. While the size of the peg cannot be estimated without additional trial and error, the strength of the robot's actions, on the other hand, can be identified after taking a handful of actions. Performing online system identification to reduce uncertainty over this parameter can allow the agent to solve the task more effectively. Thus, we propose to enhance existing robust RL solutions by introducing uncertainty set-awareness and system identification capabilities.

To this end, we formulate the multi-set robustness problem to learn a policy that is robust to multiple uncertainty sets. We then propose a framework that consists of a probabilistic system identification model and our multi-set robust policy, which we condition on the uncertainty set inferred by the model. We call our approach System Identification and Risk-Sensitive Adaptation (SIRSA). We compare SIRSA to prior methods based on robust RL and on system identification on a suite of continuous control tasks, including the 7-DoF peg insertion task in Fig.~\ref{fig:sawyer_peg}, and find substantial improvements in worst-case performance on new environments. We also find that the policy learned with SIRSA can transfer to environments with misspecified priors and with non-stationary dynamics.

\section{Related Work}
\label{sec:related_work}
Our work is at the intersection of robust control, Bayesian RL, and multi-task and meta-RL, which we review below.

\textbf{Robust and risk-sensitive RL.} The robust Markov decision process is a worst-case formulation of the RL problem with uncertainty in the transition probabilities, but can only be tractably solved in the tabular case~\citep{morimoto2000robust,nilim2005robust,iyengar2005robust,lim2013reinforcement}. Subsequent formulations treat the uncertainty as perturbations from a parameterized adversary, which can occur in the transition dynamics~\citep{pinto2017robust,mankowitz2019robust,tessler2019action,vinitsky2020robust}, the reward function~\citep{lin2020model,zahavy2020discovering}, or the underlying parameters of the environment~\cite{rajeswaran2016epopt,abraham2020model,mehta2020active}. Our work formulates a new robust control problem: robustness to a distribution over uncertainty sets. These uncertainty sets characterize uncertainty over a set of unobserved environment parameters.

Worst-case solutions can be overly pessimistic, prompting the adoption of a different risk metric, the conditional value-at-risk (CVaR), which allows control over the level of risk sensitivity through the hyperparameter $\alpha$~\citep{rockafellar2000optimization}. In RL, the CVaR objective can be optimized by sampling~\citep{tamar2015optimizing}, a distributional critic~\citep{tang2019worst}, or an ensemble of environment models~\citep{mordatch2015ensemble,rajeswaran2016epopt}. We implement a sampling-based approximation to the CVaR objective, using a learned multi-task critic.

\textbf{Bayesian RL and system identification.} Another way to handle uncertainty in RL is with the Bayes-adaptive MDP (BAMDP)~\citep{duff2002optimal,ross2007bayes} (see \citet{ghavamzadeh2016bayesian} for a review). As the agent accumulates experience, we can refine its uncertainty estimates about the environment, and adapt the policy to either the most likely MDP~\citep{yu2017preparing,yu2018policy}, a sample from the posterior over MDPs~\citep{rakelly2019efficient}, or the full belief distribution~\citep{brunskill2012bayes,guez2012efficient,guez2013scalable,lee2018bayesian,zintgraf2019varibad,abraham2020model,mozian2020learning}. 
Our work combines robust and Bayesian methods by deriving an uncertainty set from the belief and acting according to a risk-sensitive RL objective.
Another different method at this intersection is RAMCP~\citep{sharma2019robust}, which robustly plans under misspecified prior beliefs in the Bayes-adaptive MDP. A key difference in our work is that we aim to \emph{generalize} to new prior beliefs that describe novel environments. Furthermore, our experiments show that our framework can also handle misspecified priors.

While the BAMDP assumes the latent context is never observed, including at train time, we relax this assumption in our setting and access the context for each training environment to train a probabilistic system identification model via supervised learning. At test time, the model infers an uncertainty set over the true context from a partial trajectory. Unlike prior work in system identification for transfer~\citep{yu2017preparing,yu2018policy,kumar2021rma}, we address the identifiability issues in systems where different contexts cannot be distinguished, and optimize a risk-sensitive objective to act robustly with respect to the non-identifiable parameters.

\textbf{Multi-task and meta-RL.} The multi-task RL setting aims to transfer knowledge between related tasks by learning the set of tasks together~\citep{parisotto2016actor,teh2017distral,hausman2018learning,yang2020multi,yu2020gradient,sodhani2021multi}. Meta-RL is a related setting whose goal is to rapidly adapt to new tasks~\citep{finn2017model,rothfuss2018promp,nagabandi2018learning,song2020rapidly}. Notably, context-based meta-RL algorithms extract information about new tasks from a few interactions~\citep{duan2016rl,wang2016learning,perez2018efficient,rakelly2019efficient,zintgraf2018fast,lee2019stochastic}. Our method, which falls into this category, conditions the agent on the inferred uncertainty set.

\emph{Bayesian meta-RL.} Prior work in Bayesian meta-RL propose algorithms that train a policy conditioned on the posterior distribution (or belief) over the inferred context, allowing the agent to reason about task uncertainty~\citep{humplik2019meta,zintgraf2019varibad,zhang2021metacure}. However, our policy optimizes a risk-sensitive objective, rather than the expectation of the return over the belief. While Bayesian meta-RL agents balance exploration and exploitation in a new task based on the uncertainty, our work focuses less on exploration in a new task. Instead, we aim to design an agent that can robustly solve a new task under safety-critical conditions.

\emph{Robust meta-RL.} Prior work has also studied robust meta-RL but under different setups and objectives, including robustness against adversarial reward functions under a learned model~\citep{lin2020model} and robustness by learning diverse behaviors within a single MDP~\citep{kumar2020one,zahavy2020discovering}. The objective of our work is to robustly adapt to new environments by training in a set of related MDPs. Most closely related is CARL~\citep{zhang2020cautious}, which prepares the agent for safety-critical few-shot adaptation through pre-training on related source environments. CARL captures uncertainty through a probabilistic dynamics model, fine-tunes the model with new data collected in adaptation episodes with the target environment, and generates risk-sensitive plans with respect to the fine-tuned model. In contrast to CARL, which relies on multiple rounds of trial-and-error for adaptation, the robust RL setting typically evaluates \emph{zero-shot} performance on new environments. Without the opportunity to test different behaviors for adaptation, new challenges with system identifiability arise.

\section{Problem Setup}
\label{sec:problem_setup}
We first introduce notation in the standard RL setting in Sec.~\ref{subsec:prelim} and the robust contextual MDP in Sec.~\ref{subsec:rcmdp}. Then, we formalize the multi-set robustness objective in Sec.~\ref{subsec:msr}.

\newcommand{\bs}{\mathbf{s}}
\newcommand{\ba}{\mathbf{a}}
\newcommand{\cS}{\mathcal{S}}
\newcommand{\cA}{\mathcal{A}}
\newcommand{\cC}{\mathcal{C}}
\newcommand{\cM}{\mathcal{M}}
\newcommand{\cJ}{\mathcal{J}}
\newcommand{\E}{\mathbb{E}}

\subsection{Preliminaries}
\label{subsec:prelim}
A Markov decision process (MDP) or task is a tuple $\cM = \langle \cS, \cA, p, r, \rho, \gamma \rangle$ where $\cS$ is the state space, $\cA$ is the action space, $p$ is the state transition probability, $r$ is the reward function, $\rho$ is the initial state distribution, and $\gamma \in [0, 1)$ is the discount factor. The goal in standard RL is to learn a policy $\pi$ that maximizes the expected sum of rewards $\cJ(\pi) \coloneqq \E_{\pi,p} \left[ G_\pi \right]$ where $ G_\pi = \sum_{t=0}^\infty \gamma^t r(\bs_t, \ba_t)$.

We also consider the contextual Markov decision process (CMDP)~\citep{hallak2015contextual} which, like the standard MDP, is equipped with a state space $\cS$ and action space $\cA$. It additionally has a context space $\cC$ and function $\cM$ that maps any context $c \in \cC$ to an MDP $\cM(c) = \langle \cS, \cA, p^c, r^c, \rho, \gamma \rangle$, where $p^c$ and $r^c$ are parameterized by $c$.\footnote{We refer to contexts and tasks interchangeably.}

\subsection{Robust Contextual Markov Decision Process}
\label{subsec:rcmdp}

The exact context $c$ describing the task is often unknown, especially when the agent is transferred to an entirely new task. Instead, there may be a prior belief $b(c)$ over the context $c$, from which we can derive an uncertainty set.

Formally, we define the \textbf{robust contextual Markov decision process (R-CMDP)}, which extends the CMDP with an initial uncertainty set $\Xi \subseteq \cC$ over the contexts. We also assume a distribution $p(\Xi)$ from which we can draw samples. This uncertainty set is given at the beginning of an episode and can be interpreted as a prior over the true context. We focus on parameterized uncertainty sets, and $\Xi$ refers to the parameters that define the set, e.g., the center and radius for ball sets. Then, one way to acquire a robust policy is to optimize the worst-case objective with respect to the uncertainty set $\cJ_\text{min}(\pi) \coloneqq \min_{c \in \Xi} \E_{\pi,p^c} \left[ G_\pi^c \right]$ where $G_\pi^c = \sum_{t=0}^\infty \gamma^t r^c(\bs_t, \ba_t)$. In this work, we optimize a softer version of this worst-case objective: the conditional value-at-risk (CVaR)~\cite{tamar2015optimizing,rajeswaran2016epopt,tang2019worst}.

The CVaR objective is defined over the random variable $G_\pi^\Xi$ of returns induced by the uniform distribution over contexts in the uncertainty set $\Xi$. First, the value-at-risk is given by the $\alpha$-quantile of the return distribution $$\text{VaR}_\alpha (G_\pi^\Xi) \coloneqq \max \{ y | P(G_\pi^\Xi \le y) \le \alpha \}.$$ Then, denoting $\mathcal{P}(c)$ as the uniform distribution over the set $\{ c \in \Xi | G_\pi^c \le \text{VaR}_\alpha \left( G_\pi^\Xi \right) \}$, the CVaR objective is $$\cJ^{\text{CVaR}_\alpha}_\pi (\Xi) \coloneqq \E_{\pi, c\sim \mathcal{P}(c)} \left[ G_\pi^c \right],$$ the expected return over the lower $\alpha$-percentile subset of the uncertainty set. When $\alpha = 1$, the objective is the average over the perturbation set, and when $\alpha \to 0$, the objective becomes the max-min objective.

\subsection{Multi-Set Robustness}
\label{subsec:msr}
Optimizing a robust policy with respect to a new uncertainty set can be costly for each new policy. Hence, we aim to learn a single policy that can be robust to several different uncertainty sets. We do so by leveraging the multi-task RL setting to optimize a policy that can generalize to and provide good worst-case performance with respect to \emph{new} uncertainty sets.

In particular, the learner has access to $M$ training tasks $\{ \cM_i \}_{i=1}^M$ that are parameterized by $M$ different observed contexts $\{ c_i \}_{i=1}^M$. The goal of our setting is to learn a set-conditioned policy $\pi(\ba|\bs, \Xi)$ that maximizes the worst-case expected return with respect to all uncertainty sets from the distribution $p(\Xi)$. That is, we want to optimize $\E_{\Xi \sim p(\Xi)} \left[ \cJ^{\text{CVaR}_\alpha}_\pi(\Xi) \right]$.

\section{System Identification and its Challenges}
Rather than behaving invariantly to different contexts as in robust RL, another approach is to condition the policy on the context or a distribution over the context. System identification methods based on this idea train a predictive model to produce either a point estimate of or a posterior distribution over the context, given a history $h_t = (\bs_0, \ba_0, r_0, \dots, \bs_{t-1}, \ba_{t-1}, r_{t-1})$. This approach has demonstrated strong generalization performance, including transfer from simulation to the real world~\citep{yu2017preparing,yu2018policy,kumar2021rma}.

However, as discussed by~\citeauthor{dorfman2020offline}, many systems are often determined by parameters that are not easily identifiable from a limited amount of interaction. Recall our peg-insertion example from Fig.~\ref{fig:sawyer_peg}: the size of the peg cannot be determined within a single trial, but critically, the robot has to take this parameter into account to select a box to insert the peg into. In these low-data regimes, the system identification model can fail to accurately distinguish between multiple MDP contexts. Formally, the context $c$ is non-identifiable from a dataset $h$ if there is a set of other contexts $\mathcal{C}' \subseteq \mathcal{C}$ that can also explain the data. 

\begin{definition}[Context non-identifiability]
Let $P^{c, \pi}(h_{:t})$ denote the probability distribution over histories at time-step $t$ under the MDP $\mathcal{M}(c)$ and policy $\pi$. Then, the context $c$ is non-identifiable from the dataset $h_{:t}$ collected by policy $\pi$ if there exists a subset $\mathcal{C}' \not= \{ c \}$ such that $P^{c, \pi}(h_{:t}) = P^{c', \pi}(h_{:t})$ for all $c \in \mathcal{C}'$.
\end{definition}

As an aside, context non-identifiability can also be viewed as posterior collapse~\citep{wang2020posterior}. In our setting, posterior collapse occurs when the prior and posterior belief distributions are equal, i.e., $b'(c|b,h) = b(c)$. Hence, one proxy measure of context identifiability is the entropy of the belief distribution, i.e., higher entropy indicates lower identifiability of the context. This problem is further exacerbated when knowledge of the context is critical to the task at hand, i.e., confusing it with a different context can lead to a large drop in performance. 

\begin{definition}[Critical contexts]
Denote the optimal context-dependent policy by $\pi^*(c) \coloneqq \arg\max_{\pi} \E_{\pi,c}[G_\pi^c]$. Consider the set of contexts $\mathcal{C}'$ for which $P^{c, \pi^*(c)}(h) = P^{c', \pi^*(c)}(h)$ holds for all $c' \in \mathcal{C}'$. The worst-case gap for a context-dependent policy evaluated in the MDP with context $c$ is $D(c) = \max_{c' \in \mathcal{C}'} G_{\pi^*(c)}^c - G_{\pi^*(c')}^c$. 
\end{definition}

It becomes clear when there is uncertainty around a critical context $c$, the gap can be significant. Hence, to be robust to the worst case of the non-identifiable set, we define our objective to minimize the worst-case gap: $\min_\pi \max_{c \in \mathcal{C}'} G^c_{\pi^*(c)} - G^c_\pi$. In the next section, we introduce our algorithm which re-estimates the uncertainty set while taking actions that are robust at each time-step.

\section{Risk-Sensitive Adaptation via System Identification and Multi-Set Robustness}
\begin{figure*}
    \centering
    \includegraphics[width=0.85\linewidth]{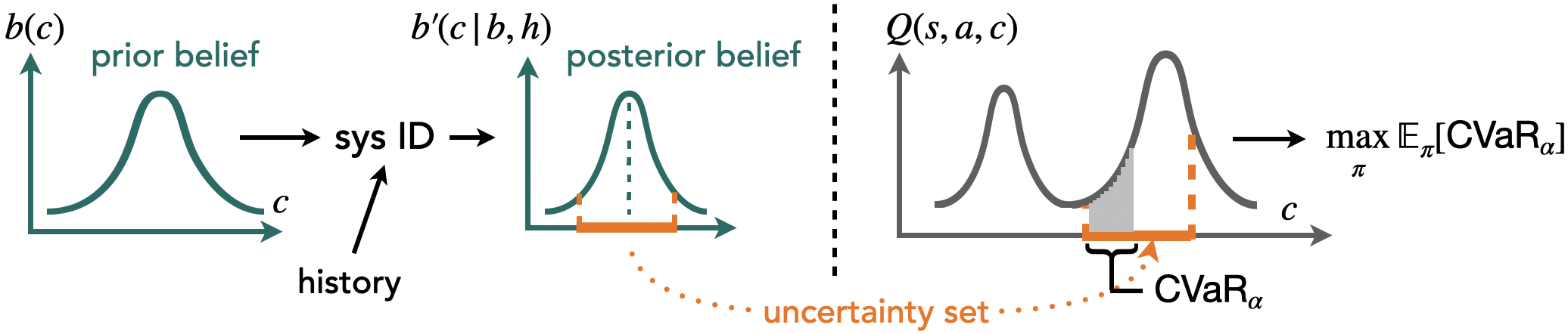}
    \vspace{-0.4cm}
    \caption{\small Our framework combines system identification with risk-sensitive RL to robustly adapt to new environments. First, the algorithm updates the uncertainty set over the context with the agent's recent history $h$. We then optimize a risk-sensitive policy over the returns within this uncertainty set.}
    \label{fig:framework}
    \vspace{-0.4cm}
\end{figure*}

To address the challenges associated with non-identifiable systems, we propose a simple framework that consists of a probabilistic system identification model and a family of risk-sensitive policies $\pi(\ba | \bs, \Xi)$ conditioned on the uncertainty set inferred by the model. The resulting algorithm combines the benefits of system identification and risk-sensitive RL as it reduces the model uncertainty where possible while behaving cautiously with respect to the irreducible uncertainty. Our overall approach, which we call System Identification and Risk-Sensitive Adaptation (SIRSA), is illustrated in Fig.~\ref{fig:framework}, with each component detailed below.

\subsection{Probabilistic System Identification}
To capture the epistemic uncertainty of an unknown environment at test time, we train an ensemble of models to predict the context $c$ that parameterizes the environment's dynamics and reward function. Recall that at train time, the agent observes the context of each training task $\{ c_i \}_{i=1}^M$, and for each task, collects a dataset of transitions $\{ (\bs_t, \ba_t, r_t, \bs'_t) \}_t$. 

We learn an ensemble of $B$ different models, where each model $f_i$ maps an initial uncertainty set $\Xi$ and a history $h$ of $H$ transitions to a context. In this work, the uncertainty set is an $\ell_1$-ball with its own center $\mu$ and width $\sigma$. Then, each model $f_i$ has parameters $\psi_i$ that are trained with the mean squared error on the predicted context:
\begin{equation} \label{eqn:ensemble}
    \mathcal{L}_{\psi_{1:B}} = \E_{(\mu, \sigma, h, c) \sim \mathcal{D}, j \sim \text{Unif}(B)} \left[ \left( f_{\psi_j} (\mu, \sigma, h) - c \right)^2 \right],
\end{equation}
where the initial uncertainty set $\Xi = (\mu, \sigma)$, given by the environment to the learner, offers an initial guess of the true context. We define the parameters of the posterior uncertainty set $\Xi' = (\mu', \sigma')$ as the mean and standard deviation of the ensemble:
\begin{equation} \label{eqn:posterior}
    \mu' = \mu\left( \left\{ f_{\psi_j}(\Xi, h) \right\}_{j=1}^B \right),~
    \sigma' = \sigma\left( \left\{ f_{\psi_j}(\Xi, h) \right\}_{j=1}^B \right),
\end{equation}
where $\mu(\cdot)$ and $\sigma(\cdot)$ compute the mean and standard deviation, respectively. 
At inference time, we recursively update the uncertainty set by using the set inferred from the previous time-step as the prior. That is, the perturbation set $\Xi_t$ at time-step $t$ has parameters $\mu_t = \mu ( \{ f_{\psi_j}(\mu_{t-1}, \sigma_{t-1}, h) \}_{j:1...B} ) $ and 
$\sigma_t = \sigma ( \{ f_{\psi_j}(\mu_{t-1}, \sigma_{t-1}, h) \}_{j:1...B} )$. Next, we describe how we optimize a risk-sensitive policy to act robustly with respect to the inferred uncertainty set.

\subsection{Risk-Sensitive Policy Optimization}

The CVaR objective is computed by finding, from the set of environments defined by the uncertainty set $\Xi$, the $\alpha$-quantile that the policy performs worst in. While the distribution of $G_\pi^\Xi$ is unknown, we can approximate its CVaR through a context-conditioned critic $Q_\theta(\bs, \ba, c)$. That is,
\begin{align*}
   \E_{c\sim \mathcal{P}(c)} \left[ G_\pi^c \right] &\approx \E_{c \sim \mathcal{P}(c)} \left[ \E_{\bs \sim \mathcal{D}, \ba \sim \pi(\cdot | \bs)} \left[ Q_\theta(\bs, \ba, c) \right] \right],
\end{align*}
where $\mathcal{P}(c)$ is the uniform distribution over the set $\{ c \in \Xi | G_\pi^c \le \text{VaR}_\alpha(G_\pi^\Xi) \}$. 

We can form a Monte-Carlo estimate of the CVaR as follows. Let $\tilde{c}_1, \dots, \tilde{c}_N$ be $N$ samples drawn i.i.d. from the uncertainty set $\Xi$, and $Q_1, \dots, Q_N$ be their corresponding Q-values, i.e., $Q_i = Q_\theta(\bs, \ba, \tilde{c}_i)$. After sorting the contexts in ascending order based on their Q-values, $\tilde{c}_{[1]}, \dots, \tilde{c}_{[N]}$, the empirical $\alpha$-quantile is simply $Q_\theta(\bs, \ba, \tilde{c}_{[\lfloor \alpha N \rfloor]})$, and, the empirical CVaR approximation is
$$
\frac{1}{\lfloor \alpha N \rfloor} \sum_{i=1}^{\lfloor \alpha N \rfloor} Q_\theta \left(\bs, \ba, \tilde{c}_{[i]} \right).
$$
To update the policy $\pi_\phi$, we can compute $\nabla_\phi J_\pi^{\text{CVaR}_\alpha}$, the gradient of the approximated CVaR with respect to the policy parameters $\phi$, with
\begin{multline} \label{eqn:cvar}
    \mathop{\mathbb{E}}_{\bs \sim \mathcal{D}} \left[ \sum_{i=1}^{\lfloor \alpha N \rfloor} \frac{\nabla_\ba Q_\theta (\bs, \ba, \tilde{c}_{[i]}) |_{\ba \sim \pi_\phi} }{\lfloor \alpha N \rfloor} \nabla_\phi \pi_\phi(\ba | \bs, \Xi) \right].
\end{multline}

\begin{algorithm}[t]
    \caption{System Identification and Risk-Sensitive Adaptation (SIRSA)}
    \label{alg:method}
    \begin{algorithmic}
    \STATE {\bfseries Input:} CVaR level $\alpha$, threshold $T_\text{threshold}$
    \STATE Initialize replay buffers for each training task $\mathcal{D}[c]$
    \FOR{$i = 1, 2, \dots$} 
        \STATE Sample training environment from set of environments
        \STATE Initialize history $h_0 = \{ \bs_0 \}$
        \FOR{each environment step}
            \STATE Take action $\ba_t \sim \pi_\phi(\cdot | \bs, \Xi_t)$
            \STATE Update history $h_{t} \gets h_{t-1} \cup \{ (\bs_t, \ba_t, r_t, \bs'_t) \}$
            \STATE Update uncertainty set $\Xi_{t+1}$ according to Eqn.~\ref{eqn:posterior}
        \ENDFOR 
        \STATE Update replay buffer $\mathcal{D}[c] \gets \mathcal{D}[c] \cup h_T$
        \FOR{each gradient step}
            \STATE Sample batch from replay buffers $\bigcup_c \mathcal{D}[c]$
            \STATE Update critic parameters $\theta$ with $\nabla_\theta \mathcal{J}_Q$
            \IF{$i < T_\text{threshold}$}
                \STATE Update actor parameters $\phi$ with $\nabla_\phi J_\pi$
            \ELSE
                \STATE Update actor parameters $\phi$ with $\nabla_\phi J^{\text{CVaR}_\alpha}_\pi$
            \ENDIF
            \STATE Update ensemble parameters $\psi_k$ with $\nabla_{\psi_k} \mathcal{L}_{\psi_k}$
        \ENDFOR
    \ENDFOR
    \end{algorithmic}
\end{algorithm}

% \subsection{Implementation Details}
We construct our CVaR actor on top of Soft Actor-Critic (SAC)~\citep{haarnoja2018soft}. Our algorithm, which we call \textbf{S}ystem \textbf{I}dentification and \textbf{R}isk-\textbf{S}ensitive \textbf{A}daptation (SIRSA), is summarized in Alg.~\ref{alg:method}. We begin by training the actor and critic with the losses $\mathcal{J}_\pi$ and $\mathcal{J}_Q$ defined in SAC, and the ensemble of models with the loss defined in Eqn.~\ref{eqn:ensemble}. After $T_\text{threshold}$ iterations, we update the actor $\pi_\phi$ based on the CVaR objective defined in Eqn.~\ref{eqn:cvar} instead of $\mathcal{J}_\pi$. Full implementation details can be found in Appendix~\ref{app:implementation}.

\section{Experiments}
\label{sec:experiments}
We design several experiments to understand the effectiveness of our proposed approach compared to system identification and robust RL approaches in unseen environments. Specifically, we seek to answer the following questions:\footnote{Code and videos of our results are on our webpage: \url{https://sites.google.com/view/sirsa-public/home}.}
\begin{enumerate}[leftmargin=*,topsep=0pt,noitemsep]
    \item How does our method SIRSA compare to standard system identification and robust RL in terms of worst-case performance on new uncertainty sets?
    
    \item Can SIRSA generalize to new test-time scenarios such as misspecified priors and non-stationary dynamics?
    
    \item How does SIRSA respond to varying $\alpha$-levels of risk sensitivity?
\end{enumerate}

\subsection{Experimental Setup}
\textbf{Baselines}. First, we consider multi-task RL baselines that train a context-conditioned policy $\pi(\ba | \bs, c)$.
\begin{itemize}[leftmargin=*,topsep=0pt,noitemsep]
    \item \textbf{Context-conditioned policy ensemble}. At \emph{test} time, $N^\text{ens}$ contexts $\{ c^i \}_{i=1}^{N^\text{ens}}$ are sampled from the initial uncertainty set to create an ensemble of policies $\pi^\text{ens}(\cdot | \bs) = \sum_{i=1}^{N^\text{ens}} \pi(\ba|\bs, c^i) / N^\text{ens}$. We use $N^\text{ens} = 5$.

    \item \textbf{Context-conditioned policy with true context (oracle)}. An oracle with access to the ground-truth context at test time, given as input to the context-conditioned policy.
\end{itemize}
We also compare to the system ID ablation of SIRSA, which optimizes the expected return rather than the CVaR:
\begin{itemize}[leftmargin=*,topsep=0pt]
    \item \textbf{Set-conditioned policy with system identification}~\citep{yu2017preparing}. Along with a set-conditioned policy $\pi(\ba | \bs, \Xi)$, this baseline trains a system identification model that maps the history of last $H$ states and actions to a belief over the context. The belief inferred by the model is given to the policy.
\end{itemize}

Finally, we compare to existing robust/risk-sensitive RL methods. Like our approach, each of these methods controls the risk level through the CVaR hyperparameter $\alpha$. We run each algorithm with $\alpha \in \{ 0.25, 0.5, 0.75, 1.0 \}$, and report the results for the most performant policy in this section. In Appendix~\ref{app:alpha}, we report the full results for each value of $\alpha$.
\begin{itemize}[leftmargin=*,topsep=0pt,noitemsep]
    \setlength{\itemsep}{0pt}
    \item \textbf{EPOpt}~\citep{rajeswaran2016epopt}. A domain randomization method that optimizes the CVaR objective by training on the $\alpha$-worst percentile of all training environments.

    \item \textbf{Multi-Set EPOpt}. We design a stronger variant of EPOpt by training a multi-set robust policy $\pi(\ba | \bs, \Xi)$ on the $\alpha$-worst percentile of environments in each set $\Xi \subseteq \mathcal{C}$. 
    
    \item \textbf{Worst Cases Policy Gradients (WCPG)}~\citep{tang2019worst}. This comparison trains a family of $\alpha$ conditional policies $\pi(\ba | \bs, \alpha)$ with varying levels of risk sensitivity. In order to approximate the CVaR across different $\alpha$-levels, the future return generated by policy $\pi$ is modeled as a Gaussian distribution and approximated by a distributional critic, allowing the CVaR to be computed in closed form. During training, we sample $\alpha$ uniformly from $[0, 1]$. At inference time, we evaluate the policy at the $\alpha$-levels $\{ 0.25, 0.5, 0.75, 1.0 \}$. Like EPOpt, this comparison trains on the entire range of contexts  as its uncertainty set.
    
    \item \textbf{Multi-Set WCPG}~\citep{tang2019worst}. We design a stronger variant of WCPG by training a multi-set robust policy $\pi(\ba | \bs, \alpha, \Xi)$ with WCPG.
\end{itemize}

\textbf{Environments}. We design several environments to evaluate our approach, and in each, vary one or more parameters that affect the dynamics and/or reward function. All methods can access the true context of each \emph{training} environment. However, at \emph{inference} time, only an initial uncertainty set is provided, and none of the methods (with the exception of the context-conditioned oracle) have access to the true parameter values. In Table~\ref{tbl:env_params}, we tabulate the ranges for the different parameters, and describe the environments below:
\begin{table}[]
\centering
\small
\begin{tabular}{lll}
    \toprule
    Environment & Uncertain Params. & Range \\
    \midrule
    \multirow{2}{*}{Point mass} & Obstacle size & $[0.025, 0.075]$ \\
                              & Velocity & $[0.06, 0.1]$ \\
    \midrule
    \multirow{3}{*}{Minitaur} & Torso mass & $[-0.2, 0.2]$ \\
                              & Leg mass & $[-0.2, 0.2]$ \\
                              & Leg failure (x4) & $[0.0, 1.0]$ \\
    \midrule
    \multirow{3}{*}{Half-cheetah}   & Torso mass & $[-0.5, 0.5]$ \\
                              & Joint friction & $[0.1, 0.9]$ \\
                              & Joint failure (x6) & $[0.0, 1.0]$ \\
    \midrule
    \multirow{2}{*}{Peg insertion} & Step size & $[0.5, 1.5]$ \\
                              & Peg size & $[0.0125, 0.0225]$ \\
    \bottomrule
\end{tabular}
\vspace{-0.2cm}
\caption{\small Range of parameter values $\mathcal{C}$ in our environments.}
\label{tbl:env_params}
\vspace{-0.5cm}
\end{table}

\begin{itemize}[leftmargin=*,topsep=0pt,noitemsep]
    \setlength{\itemsep}{0pt}
    \item \textbf{Point mass navigation}. A point mass has to navigate around a roundabout with uncertainty in the size of the roundabout and the precise velocity. We additionally design two variants: \textbf{Point mass (obstacle)} where only the obstacle size is uncertain and \textbf{Point mass (velocity)} where only the velocity of the agent is uncertain.

    \item \textbf{Minitaur}~\citep{tan2018sim}. A simulated 8-DoF minitaur robot with uncertainty in the mass and leg failure rate. 

    \item \textbf{Half-cheetah}~\citep{brockman2016openai,vinitsky2020robust}. Modified OpenAI Gym environment with uncertainty in the mass, joint friction, and joint failure rate.

    \item \textbf{Peg insertion}~\citep{zhao2020meld,schoettler2020meta}. A simulated 7-DoF Sawyer robot arm is to insert a peg into one of the boxes (see Fig.~\ref{fig:sawyer_peg}). The uncertainty is in the position controller's step size and the size of the peg.
\end{itemize}
Full descriptions of each environment are in Appendix~\ref{app:environments}.

\textbf{Evaluation metrics}. To evaluate each method, we construct test-time uncertainty sets centered around new contexts not seen during training. We then evaluate a policy's performance by, first, uniformly sampling $K$ context vectors from each set, then, rolling out the policy in each of the $K$ sampled environments. We are in particular interested in the \textbf{worst-case performance}, approximated by the minimum return of the $K$ rollouts, and additionally report the \textbf{average-case performance}, approximated by the mean return of the $K$ rollouts. In all experiments, we use $K = 50$.

\subsection{Robustness to New Uncertainty Sets}
\label{subsec:generalization}

\textbf{Point mass.}
We first seek to better understand the strengths and weaknesses of prior methods in the Point mass (obstacle) and Point mass (velocity) environments. The former represents a parameter that is difficult to precisely identify since it would require making contact with the obstacle to infer its size. In contrast, the latter parameter can be exactly estimated given a single time-step. We compare System ID to Multi-Set EPOpt, which acts as the representative of robust RL methods.
In Table~\ref{tbl:point_mass}, we compare the worst-case performance of the two methods across $20$ test uncertainty sets, and find that firstly the identification error of the obstacle size is indeed higher than that of the velocity parameter, confirming our intuition. In the Point mass (obstacle) domain, Set-EPOpt outperforms System ID as the uncertain parameter cannot be exactly identified without incurring a penalty. In the Point mass (velocity) domain, we see the reverse result: the System ID method correctly adapts to the predicted context, whereas Set-EPOpt acts conservatively without precise identification of the parameter. The trajectories taken by these policies are visualized in Appendix~\ref{app:point_mass}.

\begin{table}[]
    \centering
    \small
    \setlength{\tabcolsep}{4pt}
    \begin{tabular}{lccc}
        \toprule
        & & \multicolumn{2}{c}{Method} \\
        \cmidrule{3-4}
        Uncertain Param. & ID Error & System ID & Set-EPOpt \\
        \midrule
        Obstacle size & $0.071 \pm 0.002$ & $37.7 \pm 0.4$ & $\mathbf{39.4 \pm 0.5}$ \\
        Velocity & $0.035 \pm 0.000$ & $\mathbf{37.9 \pm 0.0}$ & $37.3 \pm 0.1$ \\
        \bottomrule
    \end{tabular}
    \vspace{-0.4cm}
    \caption{\small Worst-case performance of system ID and Set-EPOpt when (1) the obstacle size and when (2) the velocity is uncertain.}
    \label{tbl:point_mass}
    \vspace{-0.6cm}
\end{table}

\textbf{High-dimensional domains.} In Table~\ref{tbl:test_set_results}, we present the results in the remaining domains. In terms of the worst-case performance (see the ``Sample Min'' column), many of the studied baselines perform competitively against each other. System ID, which optimizes for the expectation of the return over its inferred belief, attains strong \emph{average-case} returns as a result (see the ``Sample Mean'' column). However, there is no single baseline that outperforms the rest in all settings in terms of worst-case returns, the primary metric we are interested in. On the other hand, SIRSA, which inherits from both system identification and robust RL algorithms, consistently achieves high worst-case returns across the different environments. Interestingly, all methods demonstrate similarly strong performance in the Minitaur domain, which suggests that context awareness is not as critical in this domain.
\begin{table}[ht!]
    \centering
    \small
    \setlength{\tabcolsep}{3.75pt}
    \begin{tabular}{llcc}
        \toprule
        Task & Method & Sample Min & Sample Mean \\
        \midrule
        \multirow{8}{*}{Point mass} & Ensemble & $36.8 \pm 0.3$ & $\mathbf{41.7 \pm 0.2}$ \\
                                    & System ID & $\mathbf{37.6 \pm 0.3}$ & $\mathbf{41.5 \pm 0.2}$ \\
                                    & EPOpt & $36.3 \pm 0.6$ & $40.5 \pm 0.4$ \\
                                    & Set-EPOpt & $37.1 \pm 0.5$ & $40.7 \pm 0.4$ \\
                                    & WCPG & $34.8 \pm 0.6$ & $39.3 \pm 0.5$ \\
                                    & Set-WCPG & $34.7 \pm 0.7$ & $39.0 \pm 0.5$ \\
                                    & SIRSA (Ours) & $\mathbf{37.9 \pm 0.2}$ & $\mathbf{41.7 \pm 0.1}$ \\
                                    \cmidrule{2-4}
                                    & Oracle & $38.6 \pm 0.3$ & $42.6 \pm 0.1$ \\
        \midrule
        \multirow{8}{*}{Minitaur}   & Ensemble &  $\mathbf{178.0 \pm 7.1}$ & $\mathbf{212.5 \pm 3.6}$ \\
                                    & System ID & $\mathbf{174.0 \pm 10.8}$ & $\mathbf{211.8 \pm 2.9}$ \\
                                    & EPOpt & $\mathbf{172.2 \pm 7.3}$ & $199.9 \pm 5.5$ \\
                                    & Set-EPOpt & $\mathbf{183.1 \pm 7.5}$ & $\mathbf{216.7 \pm 3.7}$ \\
                                    & WCPG & $\mathbf{165.5 \pm 17.7}$ & $\mathbf{193.7 \pm 19.6}$ \\
                                    & Set-WCPG & $\mathbf{174.5 \pm 10.0}$ & $206.9 \pm 4.7$ \\
                                    & SIRSA (Ours) & $\mathbf{187.8 \pm 7.6}$ & $\mathbf{214.3 \pm 2.5}$ \\
                                    \cmidrule{2-4}
                                    & Oracle & $172.2 \pm 4.1$ & $208.0 \pm 3.0$ \\
        \midrule
        \multirow{8}{*}{Half-cheetah} & Ensemble & $\mathbf{3988 \pm 75}$ & $4714 \pm 38$ \\
                                      & System ID & $3774 \pm 318$ & $4393 \pm 260$ \\
                                      & EPOpt  & $2272 \pm 218$ & $2717 \pm 324$  \\
                                      & Set-EPOpt & $3806 \pm 224$ & $4477 \pm 231$ \\
                                      & WCPG & $3747 \pm 229$ & $4305 \pm 314$ \\
                                      & Set-WCPG & $3871 \pm 207$ & $4433 \pm 253$ \\
                                      & SIRSA (Ours) & $\mathbf{4146 \pm 112}$ & $\mathbf{4872 \pm 74}$ \\
                                      \cmidrule{2-4}
                                      & Oracle & $4246 \pm 59$ & $4851 \pm 38$ \\
        \midrule
        \multirow{8}{*}{Peg insertion} & Ensemble & $45.2 \pm 3.0$ & $92.9 \pm 1.8$ \\
                                       & System ID & $73.7 \pm 4.3$ & $101.3 \pm 4.0$ \\
                                       & EPOpt & $43.2 \pm 5.4$ & $75.1 \pm 8.2$ \\
                                       & Set-EPOpt & $70.6 \pm 3.6$ & $96.9 \pm 2.7$ \\
                                       & WCPG & $33.8 \pm 7.2$ & $63.2 \pm 6.9$ \\
                                       & Set-WCPG & $68.3 \pm 4.6$ & $92.0 \pm 4.4$ \\
                                       & SIRSA (Ours) & $\mathbf{83.4 \pm 4.5}$ & $\mathbf{109.5 \pm 2.6}$ \\
                                       \cmidrule{2-4}
                                       & Oracle & $78.2 \pm 3.6$ & $122.0 \pm 1.5$ \\
        \bottomrule
    \end{tabular}
    \vspace{-0.3cm}
    \caption{\small We evaluate each policy on $20$ uncertainty sets at test time by drawing samples from the set and evaluating the policy's performance on each sampled environment. We report the minimum and mean performance over the samples as an approximation to the average-case and worst-case performance on the perturbation set. The means and standard errors are computed over $10$ policies for each method trained with random seeds. We bold the highest results that are more than a standard error higher than others.}
    \label{tbl:test_set_results}
    \vspace{-0.6cm}
\end{table}

In general, the multi-set robust RL baselines demonstrate better worst-case as well as average-case performance than their single-set counterparts. Without access to a more informative prior, the latter group of policies is trained to act robustly to the maximal uncertainty set: the union over all of the tasks seen during training. As a result, their behavior can be overly conservative. 

\textbf{Maximum initial uncertainty.} We now evaluate the policies learned by each method on the maximum uncertainty set in the Half-Cheetah domain, defined as the range of parameters $\mathcal{C}$ in Table~\ref{tbl:env_params}. This also represents the evaluation setup of standard single-set robust RL. We report the results in Table~\ref{tbl:nonstationary}. Notably, the single-set and multi-set robust RL methods perform comparably, where Set-EPOpt even outperforms EPOpt, indicating that multi-set robust policies can generalize even to the maximum uncertainty set $\mathcal{C}$.

\subsection{Generalization under Misspecification}
\textbf{Non-stationarity.} Most real-world environments are dynamic. For example, a robot's mass can vary over time as it carries different payloads. Does training for multi-set robustness facilitate generalization to non-stationary environments at test time? Intuitively, if the non-stationary parameters remain contained within a finite range, we can capture the changing environment with a single uncertainty set. 

\begin{figure}
    \centering
    \includegraphics[height=1.2in]{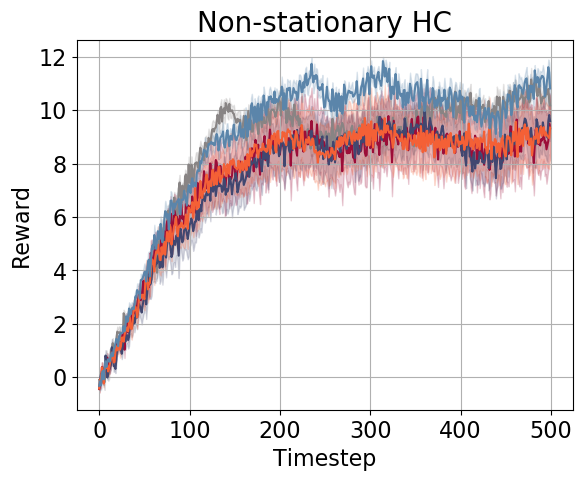}
    \includegraphics[height=1.2in]{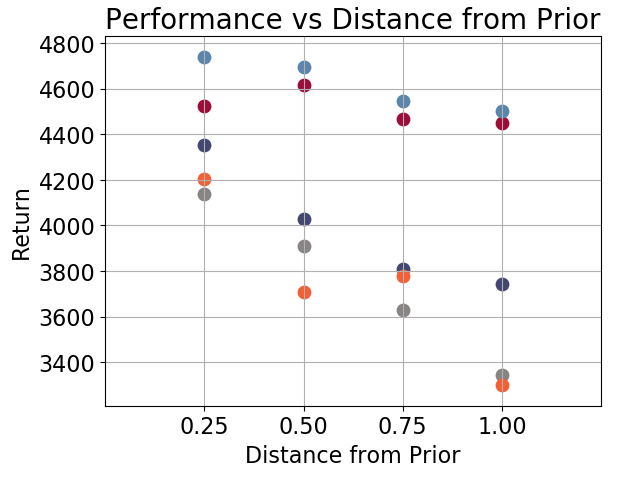}
    \includegraphics[width=\linewidth]{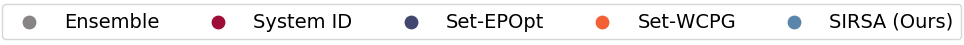}
    \vspace{-0.5cm}
    \caption{\small \textbf{Left:} Reward versus time in the Half-Cheetah environment with non-stationary parameters. The solid lines represent mean and the shaded regions represent the standard error over $10$ policies trained with different random seeds.
    \textbf{Right:} Performance on increasingly misspecified priors. Each data point is averaged over $10$ policies trained with different random seeds.}
    \label{fig:misspecification}
    \vspace{-0.2cm}
\end{figure}

In this experiment, we design a non-stationary variant of the Half-Cheetah environment. Specifically, we sample an uncertainty set at the beginning of the episode, and at every $50$ timesteps, we set the parameters of the environment to a new context sampled from the initial uncertainty set. The episode is terminated after $500$ timesteps. We report the results aggregated over $10$ different rollouts, each corresponding to a different initial uncertainty set, in Table~\ref{tbl:nonstationary}. The best-performing policy is that learned by SIRSA, indicating that adaptive risk sensitivity to a given uncertainty set is a reasonable solution to overcome non-stationarity. In Fig.~\ref{fig:misspecification} (left), we plot the reward attained by the agent over the $500$ timesteps of one of the rollouts. Here, the policy learned with SIRSA attains higher rewards at each timestep.

\textbf{Initial uncertainty set.} So far, we provided an initial uncertainty set $\Xi_0 = (\mu_0, \sigma_0)$ that correctly informs the agent about the environment. How robust is the agent when this prior is misspecified, i.e., when the initial perturbation set does not contain the true environment? In this experiment, we provide intentionally misspecified sets to the agent at test time. For each set $\Xi_0$, we sample contexts of the form $\mu_0 + w \sigma_0$, where $w \in \{ -(1+r), 1+r \}^d$, $d$ is the number of context variables, and $r$ varies between $0.25, 0.50, 0.75$, and $1.00$. We evaluate on the corresponding environments, and plot the \emph{average} over these samples as a function of $r$ in Fig.~\ref{fig:misspecification} (right). The multi-set robust RL methods, Set-EPOpt and Set-WCPG, as well as the Ensemble baseline tend to drop in performance faster as the environment deviates more from the prior uncertainty set. In contrast, SIRSA and System ID are capable of identifying contexts outside of the initial uncertainty set and degrade more gracefully.

\begin{table}[t]
    \centering
    \small
    \setlength{\tabcolsep}{10pt}
    \begin{tabular}{lcc}
        \toprule
         & Max Uncertainty & Non-stationary \\
        Method & Set Return & Env Return \\
        \midrule
        Ensemble & $3619 \pm 95$ & $4698 \pm 31$ \\
        System ID & $3608 \pm 468$ & $4277 \pm 341$ \\
        EPOpt & $1516 \pm 253$  & --- \\
        Set-EPOpt & $3360 \pm 498$ & $4336 \pm 247$ \\
        WCPG & $3598 \pm 339$ & --- \\
        Set-WCPG & $3055 \pm 456$ & $3950 \pm 486$ \\
        SIRSA (Ours) & $\mathbf{4281 \pm 73}$ & $\mathbf{4913 \pm 65}$ \\
        \midrule 
        Oracle & $4203 \pm 91$ & $4580 \pm 63$ \\
        \bottomrule
    \end{tabular}
    \vspace{-0.1cm}
    \caption{\small \textbf{Middle:} Mean and standard error of the return over $10$ random seeds on the maximum uncertainty set in the Half-Cheetah domain. \textbf{Right:} Mean and standard error of the return over $10$ random seeds on a non-stationary Half-Cheetah environment, wherein the unobserved context changes every $50$ timesteps.}
    \label{tbl:nonstationary}
\end{table}

\subsection{Sensitivity Analysis}

Our algorithm introduces several hyperparameters that determine the computation cost and the robustness of the policy. We study the CVaR hyperparameter $\alpha$ below, and study the effect of the number of CVaR samples $N$ and the effect of the system identification ensemble size $B$ in Appendix~\ref{app:sensitivity}.

\textbf{CVaR level $\alpha$.} Lower levels of $\alpha$ in principle leads to a more robust policy. However, too low levels of $\alpha$ can harm performance as the actor becomes too conservative. In terms of the computational cost, however, the algorithm computes $\lfloor \alpha N \rfloor$ gradients to approximate the CVaR gradient, 
\begin{wrapfigure}{r}{0.48\linewidth}
    \vspace{-0.5cm}
    \centering
    \includegraphics[width=\linewidth]{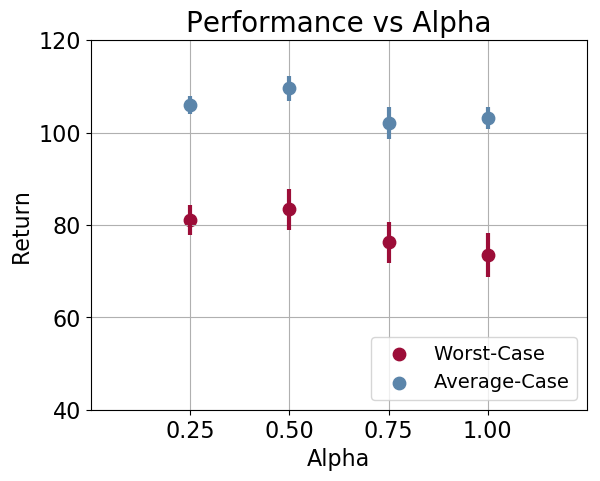}
    \vspace{-0.8cm}
    \caption{SIRSA with different values of $\alpha$.
    The plot depicts the means and standard errors over $10$ random seeds.
    }
    \label{fig:sensitivity}
\end{wrapfigure}
is therefore more efficient with smaller $\alpha$'s. In Fig.~\ref{fig:sensitivity}, we plot the average and worst-case performance of SIRSA with $\alpha \in \{ 0.25, 0.5, 0.75, 1.0 \}$ in Peg Insertion, and find that the best performing value of $\alpha$ is $0.5$, which strikes a good balance in risk sensitivity and computation cost.

\section{Discussion}
Many robust RL solutions require a prior uncertainty set over the unobserved parameters of the \emph{test-time} environment to learn a robust policy for this set at training time. To alleviate the need to build in this prior, we introduced and studied the multi-set robustness problem to facilitate generalization to new uncertainty sets. We further recognized the potential sub-optimality of memoryless robust RL methods in systems with parameters that can be identified from a history of interactions, and designed a framework that combines probabilistic system identification with the multi-set robust RL objective. Our method improves upon existing methods on a range of challenging control domains in terms of worst-case performance on new uncertainty sets.

While we believe the multi-set robustness problem represents a more general and useful framing of robustness to variations, there is also a number of interesting future directions. For example, SIRSA currently assumes the contexts that underlie the training tasks are observed to train an ensemble of predictive models via supervised learning. To remove this assumption, one can leverage tools from unsupervised representation learning to learn a representation of the true context. Another question is whether there are robustness benefits when the agent explicitly seeks exploratory actions that minimize its uncertainty over the parameters. In our experiments, we designed two types of parameters: non-identifiable parameters whose uncertainty cannot be reduced at all and identifiable parameters whose uncertainty can be reduced within a single timestep. In settings where identifiable parameters require coordinated sequences of actions to reduce uncertainty over, the agent needs to be able to balance exploration, exploitation, and robustness.

\newpage
\bibliography{references}

\begin{thebibliography}{59}
\providecommand{\natexlab}[1]{#1}
\providecommand{\url}[1]{\texttt{#1}}
\expandafter\ifx\csname urlstyle\endcsname\relax
  \providecommand{\doi}[1]{doi: #1}\else
  \providecommand{\doi}{doi: \begingroup \urlstyle{rm}\Url}\fi

\bibitem[Abraham et~al.(2020)Abraham, Handa, Ratliff, Lowrey, Murphey, and
  Fox]{abraham2020model}
Abraham, I., Handa, A., Ratliff, N., Lowrey, K., Murphey, T.~D., and Fox, D.
\newblock Model-based generalization under parameter uncertainty using path
  integral control.
\newblock \emph{IEEE Robotics and Automation Letters}, 5\penalty0 (2):\penalty0
  2864--2871, 2020.

\bibitem[Brockman et~al.(2016)Brockman, Cheung, Pettersson, Schneider,
  Schulman, Tang, and Zaremba]{brockman2016openai}
Brockman, G., Cheung, V., Pettersson, L., Schneider, J., Schulman, J., Tang,
  J., and Zaremba, W.
\newblock Openai gym.
\newblock \emph{arXiv preprint arXiv:1606.01540}, 2016.

\bibitem[Brunskill(2012)]{brunskill2012bayes}
Brunskill, E.
\newblock Bayes-optimal reinforcement learning for discrete uncertainty
  domains.
\newblock In \emph{Proceedings of the 11th International Conference on
  Autonomous Agents and Multiagent Systems-Volume 3}, pp.\  1385--1386, 2012.

\bibitem[Chen et~al.(2021)Chen, Wang, Zhou, and Ross]{chen2021randomized}
Chen, X., Wang, C., Zhou, Z., and Ross, K.
\newblock Randomized ensembled double q-learning: Learning fast without a
  model.
\newblock \emph{International Conference on Learning Representations (ICLR)},
  2021.

\bibitem[Dorfman \& Tamar(2021)Dorfman and Tamar]{dorfman2020offline}
Dorfman, R. and Tamar, A.
\newblock Offline meta reinforcement learning -- identifiability challenges and
  effective data collection strategies.
\newblock \emph{Neural Information Processing Systems (NeurIPS)}, 2021.

\bibitem[Duan et~al.(2016)Duan, Schulman, Chen, Bartlett, Sutskever, and
  Abbeel]{duan2016rl}
Duan, Y., Schulman, J., Chen, X., Bartlett, P.~L., Sutskever, I., and Abbeel,
  P.
\newblock {RL2}: Fast reinforcement learning via slow reinforcement learning.
\newblock \emph{arXiv preprint arXiv:1611.02779}, 2016.

\bibitem[Duff(2002)]{duff2002optimal}
Duff, M.~O.
\newblock \emph{Optimal Learning: Computational procedures for Bayes-adaptive
  Markov decision processes}.
\newblock University of Massachusetts Amherst, 2002.

\bibitem[Finn et~al.(2017)Finn, Abbeel, and Levine]{finn2017model}
Finn, C., Abbeel, P., and Levine, S.
\newblock Model-agnostic meta-learning for fast adaptation of deep networks.
\newblock \emph{International Conference on Machine Learning (ICML)}, 2017.

\bibitem[Ghavamzadeh et~al.(2016)Ghavamzadeh, Mannor, Pineau, and
  Tamar]{ghavamzadeh2016bayesian}
Ghavamzadeh, M., Mannor, S., Pineau, J., and Tamar, A.
\newblock Bayesian reinforcement learning: A survey.
\newblock \emph{arXiv preprint arXiv:1609.04436}, 2016.

\bibitem[Guez et~al.(2012)Guez, Silver, and Dayan]{guez2012efficient}
Guez, A., Silver, D., and Dayan, P.
\newblock Efficient bayes-adaptive reinforcement learning using sample-based
  search.
\newblock \emph{Neural Information Processing Systems (NeurIPS)}, 2012.

\bibitem[Guez et~al.(2013)Guez, Silver, and Dayan]{guez2013scalable}
Guez, A., Silver, D., and Dayan, P.
\newblock Scalable and efficient bayes-adaptive reinforcement learning based on
  monte-carlo tree search.
\newblock \emph{Journal of Artificial Intelligence Research}, 48:\penalty0
  841--883, 2013.

\bibitem[Haarnoja et~al.(2018)Haarnoja, Zhou, Abbeel, and
  Levine]{haarnoja2018soft}
Haarnoja, T., Zhou, A., Abbeel, P., and Levine, S.
\newblock Soft actor-critic: Off-policy maximum entropy deep reinforcement
  learning with a stochastic actor.
\newblock In \emph{International conference on machine learning}, pp.\
  1861--1870. PMLR, 2018.

\bibitem[Hallak et~al.(2015)Hallak, Di~Castro, and
  Mannor]{hallak2015contextual}
Hallak, A., Di~Castro, D., and Mannor, S.
\newblock Contextual markov decision processes.
\newblock \emph{arXiv preprint arXiv:1502.02259}, 2015.

\bibitem[Hausman et~al.(2018)Hausman, Springenberg, Wang, Heess, and
  Riedmiller]{hausman2018learning}
Hausman, K., Springenberg, J.~T., Wang, Z., Heess, N., and Riedmiller, M.
\newblock Learning an embedding space for transferable robot skills.
\newblock \emph{International Conference on Learning Representations (ICLR)},
  2018.

\bibitem[Humplik et~al.(2019)Humplik, Galashov, Hasenclever, Ortega, Teh, and
  Heess]{humplik2019meta}
Humplik, J., Galashov, A., Hasenclever, L., Ortega, P.~A., Teh, Y.~W., and
  Heess, N.
\newblock Meta reinforcement learning as task inference.
\newblock \emph{arXiv preprint arXiv:1905.06424}, 2019.

\bibitem[Iyengar(2005)]{iyengar2005robust}
Iyengar, G.~N.
\newblock Robust dynamic programming.
\newblock \emph{Mathematics of Operations Research}, 30\penalty0 (2):\penalty0
  257--280, 2005.

\bibitem[Kumar et~al.(2021)Kumar, Fu, Pathak, and Malik]{kumar2021rma}
Kumar, A., Fu, Z., Pathak, D., and Malik, J.
\newblock Rma: Rapid motor adaptation for legged robots.
\newblock \emph{Robotics: Science and Systems (RSS)}, 2021.

\bibitem[Kumar et~al.(2020)Kumar, Kumar, Levine, and Finn]{kumar2020one}
Kumar, S., Kumar, A., Levine, S., and Finn, C.
\newblock One solution is not all you need: Few-shot extrapolation via
  structured maxent rl.
\newblock \emph{Neural Information Processing Systems (NeurIPS)}, 33, 2020.

\bibitem[Lee et~al.(2020)Lee, Nagabandi, Abbeel, and Levine]{lee2019stochastic}
Lee, A.~X., Nagabandi, A., Abbeel, P., and Levine, S.
\newblock Stochastic latent actor-critic: Deep reinforcement learning with a
  latent variable model.
\newblock \emph{Neural Information Processing Systems (NeurIPS)}, 2020.

\bibitem[Lee et~al.(2019)Lee, Hou, Mandalika, Lee, Choudhury, and
  Srinivasa]{lee2018bayesian}
Lee, G., Hou, B., Mandalika, A., Lee, J., Choudhury, S., and Srinivasa, S.~S.
\newblock Bayesian policy optimization for model uncertainty.
\newblock \emph{International Conference on Learning Representations (ICLR)},
  2019.

\bibitem[Lim et~al.(2013)Lim, Xu, and Mannor]{lim2013reinforcement}
Lim, S.~H., Xu, H., and Mannor, S.
\newblock Reinforcement learning in robust markov decision processes.
\newblock \emph{Neural Information Processing Systems (NeurIPS)}, 26:\penalty0
  701--709, 2013.

\bibitem[Lin et~al.(2020)Lin, Thomas, Yang, and Ma]{lin2020model}
Lin, Z., Thomas, G., Yang, G., and Ma, T.
\newblock Model-based adversarial meta-reinforcement learning.
\newblock In \emph{Neural Information Processing Systems (NeurIPS)}, 2020.

\bibitem[Mankowitz et~al.(2019)Mankowitz, Levine, Jeong, Abdolmaleki,
  Springenberg, Shi, Kay, Hester, Mann, and Riedmiller]{mankowitz2019robust}
Mankowitz, D.~J., Levine, N., Jeong, R., Abdolmaleki, A., Springenberg, J.~T.,
  Shi, Y., Kay, J., Hester, T., Mann, T., and Riedmiller, M.
\newblock Robust reinforcement learning for continuous control with model
  misspecification.
\newblock In \emph{International Conference on Learning Representations
  (ICLR)}, 2019.

\bibitem[Mehta et~al.(2020)Mehta, Diaz, Golemo, Pal, and
  Paull]{mehta2020active}
Mehta, B., Diaz, M., Golemo, F., Pal, C.~J., and Paull, L.
\newblock Active domain randomization.
\newblock In \emph{Conference on Robot Learning (CoRL)}, pp.\  1162--1176.
  PMLR, 2020.

\bibitem[Mordatch et~al.(2015)Mordatch, Lowrey, and
  Todorov]{mordatch2015ensemble}
Mordatch, I., Lowrey, K., and Todorov, E.
\newblock Ensemble-cio: Full-body dynamic motion planning that transfers to
  physical humanoids.
\newblock In \emph{2015 IEEE/RSJ International Conference on Intelligent Robots
  and Systems (IROS)}, pp.\  5307--5314. IEEE, 2015.

\bibitem[Morimoto \& Doya(2000)Morimoto and Doya]{morimoto2000robust}
Morimoto, J. and Doya, K.
\newblock Robust reinforcement learning.
\newblock \emph{Neural Information Processing Systems (NeurIPS)}, pp.\
  1061--1067, 2000.

\bibitem[Mozian et~al.(2020)Mozian, Higuera, Meger, and
  Dudek]{mozian2020learning}
Mozian, M., Higuera, J. C.~G., Meger, D., and Dudek, G.
\newblock Learning domain randomization distributions for training robust
  locomotion policies.
\newblock In \emph{2020 IEEE/RSJ International Conference on Intelligent Robots
  and Systems (IROS)}, pp.\  6112--6117. IEEE, 2020.

\bibitem[Nagabandi et~al.(2019)Nagabandi, Clavera, Liu, Fearing, Abbeel,
  Levine, and Finn]{nagabandi2018learning}
Nagabandi, A., Clavera, I., Liu, S., Fearing, R.~S., Abbeel, P., Levine, S.,
  and Finn, C.
\newblock Learning to adapt in dynamic, real-world environments through
  meta-reinforcement learning.
\newblock \emph{International Conference on Learning Representations (ICLR)},
  2019.

\bibitem[Nilim \& El~Ghaoui(2005)Nilim and El~Ghaoui]{nilim2005robust}
Nilim, A. and El~Ghaoui, L.
\newblock Robust control of markov decision processes with uncertain transition
  matrices.
\newblock \emph{Operations Research}, 53\penalty0 (5):\penalty0 780--798, 2005.

\bibitem[Parisotto et~al.(2016)Parisotto, Ba, and
  Salakhutdinov]{parisotto2016actor}
Parisotto, E., Ba, L.~J., and Salakhutdinov, R.
\newblock Actor-mimic: Deep multitask and transfer reinforcement learning.
\newblock \emph{International Conference on Learning Representations (ICLR)},
  2016.

\bibitem[Perez et~al.(2018)Perez, Such, and Karaletsos]{perez2018efficient}
Perez, C.~F., Such, F.~P., and Karaletsos, T.
\newblock Efficient transfer learning and online adaptation with latent
  variable models for continuous control.
\newblock \emph{arXiv preprint arXiv:1812.03399}, 2018.

\bibitem[Pinto et~al.(2017)Pinto, Davidson, Sukthankar, and
  Gupta]{pinto2017robust}
Pinto, L., Davidson, J., Sukthankar, R., and Gupta, A.
\newblock Robust adversarial reinforcement learning.
\newblock In \emph{International Conference on Machine Learning (ICML)}, pp.\
  2817--2826. PMLR, 2017.

\bibitem[Rajeswaran et~al.(2016)Rajeswaran, Ghotra, Ravindran, and
  Levine]{rajeswaran2016epopt}
Rajeswaran, A., Ghotra, S., Ravindran, B., and Levine, S.
\newblock Epopt: Learning robust neural network policies using model ensembles.
\newblock In \emph{International Conference on Learning Representations
  (ICLR)}, 2016.

\bibitem[Rakelly et~al.(2019)Rakelly, Zhou, Quillen, Finn, and
  Levine]{rakelly2019efficient}
Rakelly, K., Zhou, A., Quillen, D., Finn, C., and Levine, S.
\newblock Efficient off-policy meta-reinforcement learning via probabilistic
  context variables.
\newblock \emph{International Conference on Machine Learning (ICML)}, 2019.

\bibitem[Rockafellar et~al.(2000)Rockafellar, Uryasev,
  et~al.]{rockafellar2000optimization}
Rockafellar, R.~T., Uryasev, S., et~al.
\newblock Optimization of conditional value-at-risk.
\newblock \emph{Journal of risk}, 2:\penalty0 21--42, 2000.

\bibitem[Ross et~al.(2007)Ross, Chaib-draa, and Pineau]{ross2007bayes}
Ross, S., Chaib-draa, B., and Pineau, J.
\newblock Bayes-adaptive pomdps.
\newblock \emph{Neural Information Processing Systems (NeurIPS)}, pp.\
  1225--1232, 2007.

\bibitem[Rothfuss et~al.(2019)Rothfuss, Lee, Clavera, Asfour, and
  Abbeel]{rothfuss2018promp}
Rothfuss, J., Lee, D., Clavera, I., Asfour, T., and Abbeel, P.
\newblock Promp: Proximal meta-policy search.
\newblock \emph{International Conference on Learning Representations (ICLR)},
  2019.

\bibitem[Schoettler et~al.(2020)Schoettler, Nair, Ojea, Levine, and
  Solowjow]{schoettler2020meta}
Schoettler, G., Nair, A., Ojea, J.~A., Levine, S., and Solowjow, E.
\newblock Meta-reinforcement learning for robotic industrial insertion tasks.
\newblock In \emph{2020 IEEE/RSJ International Conference on Intelligent Robots
  and Systems (IROS)}, pp.\  9728--9735. IEEE, 2020.

\bibitem[Sharma et~al.(2019)Sharma, Harrison, Tsao, and
  Pavone]{sharma2019robust}
Sharma, A., Harrison, J., Tsao, M., and Pavone, M.
\newblock Robust and adaptive planning under model uncertainty.
\newblock In \emph{Proceedings of the International Conference on Automated
  Planning and Scheduling}, volume~29, pp.\  410--418, 2019.

\bibitem[Sodhani et~al.(2021)Sodhani, Zhang, and Pineau]{sodhani2021multi}
Sodhani, S., Zhang, A., and Pineau, J.
\newblock Multi-task reinforcement learning with context-based representations.
\newblock \emph{International Conference on Machine Learning (ICML)}, 2021.

\bibitem[Song et~al.(2020)Song, Yang, Choromanski, Caluwaerts, Gao, Finn, and
  Tan]{song2020rapidly}
Song, X., Yang, Y., Choromanski, K., Caluwaerts, K., Gao, W., Finn, C., and
  Tan, J.
\newblock Rapidly adaptable legged robots via evolutionary meta-learning.
\newblock In \emph{2020 IEEE/RSJ International Conference on Intelligent Robots
  and Systems (IROS)}, pp.\  3769--3776. IEEE, 2020.

\bibitem[Tamar et~al.(2015)Tamar, Glassner, and Mannor]{tamar2015optimizing}
Tamar, A., Glassner, Y., and Mannor, S.
\newblock Optimizing the cvar via sampling.
\newblock In \emph{Twenty-Ninth AAAI Conference on Artificial Intelligence},
  2015.

\bibitem[Tan et~al.(2018)Tan, Zhang, Coumans, Iscen, Bai, Hafner, Bohez, and
  Vanhoucke]{tan2018sim}
Tan, J., Zhang, T., Coumans, E., Iscen, A., Bai, Y., Hafner, D., Bohez, S., and
  Vanhoucke, V.
\newblock Sim-to-real: Learning agile locomotion for quadruped robots.
\newblock \emph{Robotics: Science and Systems (RSS)}, 2018.

\bibitem[Tang et~al.(2019)Tang, Zhang, and Salakhutdinov]{tang2019worst}
Tang, Y.~C., Zhang, J., and Salakhutdinov, R.
\newblock Worst cases policy gradients.
\newblock \emph{Conference on Robot Learning (CoRL)}, 2019.

\bibitem[Teh et~al.(2017)Teh, Bapst, Czarnecki, Quan, Kirkpatrick, Hadsell,
  Heess, and Pascanu]{teh2017distral}
Teh, Y.~W., Bapst, V., Czarnecki, W.~M., Quan, J., Kirkpatrick, J., Hadsell,
  R., Heess, N., and Pascanu, R.
\newblock Distral: Robust multitask reinforcement learning.
\newblock \emph{Neural Information Processing Systems (NeurIPS)}, 2017.

\bibitem[Tessler et~al.(2019)Tessler, Efroni, and Mannor]{tessler2019action}
Tessler, C., Efroni, Y., and Mannor, S.
\newblock Action robust reinforcement learning and applications in continuous
  control.
\newblock In \emph{International Conference on Machine Learning (ICML)}, pp.\
  6215--6224. PMLR, 2019.

\bibitem[Vinitsky et~al.(2020)Vinitsky, Du, Parvate, Jang, Abbeel, and
  Bayen]{vinitsky2020robust}
Vinitsky, E., Du, Y., Parvate, K., Jang, K., Abbeel, P., and Bayen, A.
\newblock Robust reinforcement learning using adversarial populations.
\newblock \emph{arXiv preprint arXiv:2008.01825}, 2020.

\bibitem[Wang et~al.(2016)Wang, Kurth-Nelson, Tirumala, Soyer, Leibo, Munos,
  Blundell, Kumaran, and Botvinick]{wang2016learning}
Wang, J.~X., Kurth-Nelson, Z., Tirumala, D., Soyer, H., Leibo, J.~Z., Munos,
  R., Blundell, C., Kumaran, D., and Botvinick, M.
\newblock Learning to reinforcement learn.
\newblock \emph{arXiv preprint arXiv:1611.05763}, 2016.

\bibitem[Wang \& Cunningham(2020)Wang and Cunningham]{wang2020posterior}
Wang, Y. and Cunningham, J.~P.
\newblock Posterior collapse and latent variable non-identifiability.
\newblock In \emph{Third Symposium on Advances in Approximate Bayesian
  Inference}, 2020.

\bibitem[Yang et~al.(2020)Yang, Xu, Wu, and Wang]{yang2020multi}
Yang, R., Xu, H., Wu, Y., and Wang, X.
\newblock Multi-task reinforcement learning with soft modularization.
\newblock \emph{Neural Information Processing Systems (NeurIPS)}, 2020.

\bibitem[Yu et~al.(2020)Yu, Kumar, Gupta, Levine, Hausman, and
  Finn]{yu2020gradient}
Yu, T., Kumar, S., Gupta, A., Levine, S., Hausman, K., and Finn, C.
\newblock Gradient surgery for multi-task learning.
\newblock \emph{Neural Information Processing Systems (NeurIPS)}, 2020.

\bibitem[Yu et~al.(2017)Yu, Tan, Liu, and Turk]{yu2017preparing}
Yu, W., Tan, J., Liu, C.~K., and Turk, G.
\newblock Preparing for the unknown: Learning a universal policy with online
  system identification.
\newblock \emph{Robotics: Science and Systems (RSS)}, 2017.

\bibitem[Yu et~al.(2019)Yu, Liu, and Turk]{yu2018policy}
Yu, W., Liu, C.~K., and Turk, G.
\newblock Policy transfer with strategy optimization.
\newblock \emph{International Conference on Learning Representations (ICLR)},
  2019.

\bibitem[Zahavy et~al.(2020)Zahavy, Barreto, Mankowitz, Hou, O'Donoghue,
  Kemaev, and Singh]{zahavy2020discovering}
Zahavy, T., Barreto, A., Mankowitz, D.~J., Hou, S., O'Donoghue, B., Kemaev, I.,
  and Singh, S.
\newblock Discovering a set of policies for the worst case reward.
\newblock In \emph{International Conference on Learning Representations
  (ICLR)}, 2020.

\bibitem[Zhang et~al.(2020)Zhang, Cheung, Finn, Levine, and
  Jayaraman]{zhang2020cautious}
Zhang, J., Cheung, B., Finn, C., Levine, S., and Jayaraman, D.
\newblock Cautious adaptation for reinforcement learning in safety-critical
  settings.
\newblock In \emph{International Conference on Machine Learning}, pp.\
  11055--11065. PMLR, 2020.

\bibitem[Zhang et~al.(2021)Zhang, Wang, Hu, Chen, Chen, Fan, and
  Zhang]{zhang2021metacure}
Zhang, J., Wang, J., Hu, H., Chen, T., Chen, Y., Fan, C., and Zhang, C.
\newblock Metacure: Meta reinforcement learning with empowerment-driven
  exploration.
\newblock In \emph{International Conference on Machine Learning}, pp.\
  12600--12610. PMLR, 2021.

\bibitem[Zhao et~al.(2020)Zhao, Nagabandi, Rakelly, Finn, and
  Levine]{zhao2020meld}
Zhao, T.~Z., Nagabandi, A., Rakelly, K., Finn, C., and Levine, S.
\newblock Meld: Meta-reinforcement learning from images via latent state
  models.
\newblock \emph{Conference on Robot Learning (CoRL)}, 2020.

\bibitem[Zintgraf et~al.(2020)Zintgraf, Shiarlis, Igl, Schulze, Gal, Hofmann,
  and Whiteson]{zintgraf2019varibad}
Zintgraf, L., Shiarlis, K., Igl, M., Schulze, S., Gal, Y., Hofmann, K., and
  Whiteson, S.
\newblock Varibad: A very good method for bayes-adaptive deep rl via
  meta-learning.
\newblock \emph{International Conference on Learning Representations (ICLR)},
  2020.

\bibitem[Zintgraf et~al.(2019)Zintgraf, Shiarlis, Kurin, Hofmann, and
  Whiteson]{zintgraf2018fast}
Zintgraf, L.~M., Shiarlis, K., Kurin, V., Hofmann, K., and Whiteson, S.
\newblock Fast context adaptation via meta-learning.
\newblock \emph{International Conference on Machine Learning (ICML)}, 2019.

\end{thebibliography}
\bibliographystyle{icml2022}

\clearpage
\newpage

\section*{Appendix}
\renewcommand{\thesubsection}{\Alph{subsection}}

\subsection{Implementation Details}
\label{app:implementation}
Below, we provide details of the implementation of our algorithm SIRSA and the baselines.

\subsubsection{SIRSA (Ours)}
The System ID baseline shares the same implementation details and hyperparameters as SIRSA, except it does not implement the CVaR objective. 

\textbf{System identification model.} We train an ensemble of $B = 4$ models, which are MLPs with $2$ fully-connected layers of size $64$ in the Point Mass domain; $2$ fully-connected laters of size $256$ in all other domains. Each model takes a $(\bs, \ba, \bs')$ tuple, outputs a prediction for the context, and is trained with the MSE of the predicted and true context.

\textbf{Policy and critic networks.} The policy and critic networks are MLPs with $2$ fully-connected layers of size $64$ in the Point Mass domain; $2$ fully-connected layers of size $256$ in all other domains. 

In the Minitaur domain, training the critic was somewhat unstable. We therefore implement REDQ~\citep{chen2021randomized}, which has been found to stabilize and accelerate learning. It trains an ensemble of $M$ critic networks. To compute the Q-values, REDQ randomly subsamples $2$ of the critic networks and take their minimum. In our Minitaur experiments, we use $M = 8$ \textbf{for our method as well as all comparisons}.

\textbf{CVaR approximation.} In our experiments, we use $N = 50$ CVaR samples to approximate the gradient of the CVaR.

\textbf{Training phases.} 
Before updating the policy with the CVaR objective, we first train the actor and critic networks with the SAC objectives:
$$
\mathcal{J}_Q = \E_{(\bs, \ba)\sim \mathcal{D}} \left[ \frac{1}{2} \left( Q_\theta (\bs, \ba) - \hat{Q}(\bs, \ba) \right)^2 \right]
$$
with 
$$
\hat{Q}(\bs, \ba) = r(\bs, \ba) + \gamma \E_{s' \sim p}\left[ V_\psi (s') \right]
$$
where $V_\psi$ is a target network, whose weights are an exponentially moving average of the value function weights.
Then, after $T_\text{threshold}$ iterations, we optimize policy with the CVaR objective defined in Eqn.~\ref{eqn:cvar} instead. 

In Point Mass, we optimize the SAC objectives for $25$K iterations then optimize the CVaR for another $25$K iterations, for a total of $50$K training iterations. In the Minitaur and Peg Insertion domains, we pre-train for $150$K iterations then optimize CVaR for $150$K iterations for a total of $300$K. In Half-Cheetah, the pre-training is $2.5$M, and the the CVaR optimization is $0.5$M long, for a total of $3$M steps.

\subsubsection{EPOpt~\citep{rajeswaran2016epopt}}
\textbf{Policy and critic networks.} The policy and critic networks are MLPs with $2$ fully-connected layers of size $64$ in the Point Mass domain, and with $2$ fully-connected layers of size $256$ in all other domains. These networks are trained with the SAC objectives. To sample a batch of size $D$ that they train on, we first sample $D / \alpha$ s-a-s' tuples from the replay buffer. Then, we sort the samples by the return of the trajectory they came from, and return the lowest $D$ tuples.

\subsubsection{WCPG~\citep{tang2019worst}}
\textbf{Policy and critic networks.} The policy and critic networks are MLPs with $2$ fully-connected layers of size $64$ in the Point Mass domain; $2$ fully-connected layers of size $256$ in all other domains. These networks are trained with SAC. 

\textbf{Q-variance network.} WCPG requires an estimate of the variance of $Q(\bs, \ba, \Xi)$. We train an MLP with $2$ fully-connected layers of size $256$ via the MSE to predict the variance of $Q(\bs, \ba, \Xi)$. To generate the target variance this network regresses to, we generate a Monte-Carlo approximation: we sample $50$ contexts $\{ c_i \}_{i:1...50}$ from $\Xi$, evaluate them with our context-conditioned critic $Q_\theta(\bs, \ba, c_i)$, and compute their sample variance. 

\textbf{CVaR approximation.} WCPG assumes that the Q-values follow a Gaussian distribution, and can therefore compute the CVaR of the uncertainty set $\Xi = (\mu, \sigma)$ in closed form as follows:
$$
Q(\bs, \ba, \mu) - (\phi(\alpha) / \Phi(\alpha)) \sqrt{\text{Q-Var}(\bs, \ba, \Xi)},
$$
where $\text{Q-Var}(\bs, \ba, \Xi)$ represents the output of the Q-variance network, $\phi(\cdot)$ is the standard normal distribution, and $\Phi(\cdot)$ is its CDF:
$$
\Psi(x) = \frac{1}{2}(1 + \text{erf}(x/\sqrt{2})).
$$

\subsection{Environment Details}
\label{app:environments}
\begin{figure*}
    \centering
    \includegraphics[width=\linewidth]{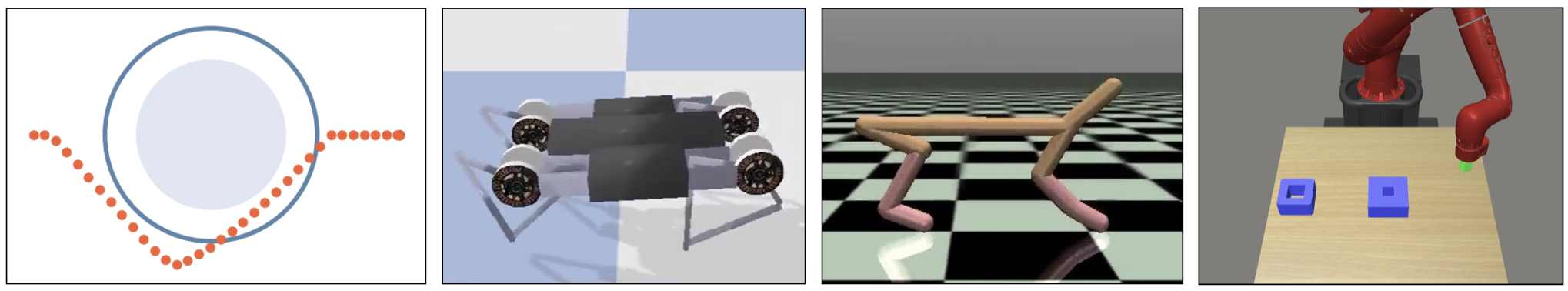}
    \caption{The domains in our evaluation. \textbf{Point Mass:} In this domain, there is uncertainty in the size of the obstacle (in lavender). The blue concentric circle represents the worst-case obstacle size within the uncertainty set. The agent must navigate around the obstacle. Its path is highlighted in pink-fuschia. \textbf{Minitaur:} The uncertainty lies in the mass of the robot and the failure rate of one of its legs. \textbf{Half-Cheetah:} The uncertainty lies in the mass of the agent, the friction of its joints, and failure rate of one of its joints. \textbf{Peg Insertion:} This domain has uncertainty over the size of the peg and the step size of the robot's actions.}
    \label{fig:envs}
\end{figure*}

In this section, we provide details of each of the four experimental domains. The domains are visualized in Fig.~\ref{fig:envs}.

\subsubsection{Point Mass}
In this environment, the agent is a point mass particle and must go around the roundabout to the goal on its other side. The $x$-velocity of the agent is fixed within a task but its precise value is unknown. The size of the roundabout $r$ is also unknown. Each episode is $50$ timesteps long, and the state consists of the $xy$-position of the agent and whether the agent is on top of the roundabout. There is one action input which controls the change in $y$-position. The reward function is defined as
$$
r_t = 1 - \mathbbm{1}(x_t^2 + y_t^2 < r^2) - 8|y_t|.
$$
Hence, the agent is encouraged to take a tight turn around the roundabout without colliding with it. We train on $20$ different uncertainty sets, with $3$ sampled contexts from each set, for a total of $60$ different contexts.

\subsubsection{Minitaur}
This environment simulates an $8$-DoF minitaur robot whose objective is to walk forward as quickly as possible. The mass of the robot's base and mass of the legs vary across tasks but are unknown. In each task, there is also a probability of failure $p_\text{fail}$ for one of the four legs, which is also unknown. Each episode terminates when the robot falls or after $500$ timesteps. At each timestep, the action input to one leg is dropped with probability $p_\text{fail}$. The agent's state consists of the robot's roll, roll rate, pitch, pitch rate, and the angles of each of the eight motors. We also append the history of last five actions, which the reward function depends on. The reward function is defined as
$$
r_t = v_t - 0.01 \| \ba_t - 2 \ba_{t-1} + \ba_{t-2} \|
$$
where $v_t$ is the observed velocity of the robot. We train on $80$ different uncertainty sets, with $10$ sampled contexts from each set, for a total of $800$ different contexts.

\subsubsection{Half-Cheetah}
This environment modifies the Half-Cheetah task from OpenAI Gym~\citep{brockman2016openai}. The agent's objective is to run forward as quickly as possible, starting from rest. The mass of the agent and the friction of the joints vary across tasks and are unknown. Like in the Minitaur environment, there is a probability of failure $p_\text{fail}$ for one of the six joints, which varies across tasks. Each episode lasts $500$ timesteps. At each timestep, the action input to one of the joints is dropped with probability $p_\text{fail}$. The agent's state consists of the velocity of the agent's center of mass and angular velocity of each of its six joints, and actions correspond to torques applied to each joint. The reward function is
$$
r_t = v_t - 0.05 \|\ba_t\|
$$
where $v_t$ is the observed velocity of the agent. We train on $80$ different uncertainty sets, with $10$ sampled contexts from each set, for a total of $800$ different contexts.

\subsubsection{Sawyer Peg Insertion}
In this modified peg-insertion task~\citep{zhao2020meld}, a $7$-DoF Sawyer robot arm needs to insert the peg attached to its end-effector into one of the two boxes in as few timesteps as possible. Across tasks, the scaling of the joint position controller and size of the peg vary and are unknown. The first box is closer to the initial position of the robot, but it also has a smaller hole. The second box on the other hand is farther from the initial position, and has a larger hole that will allow the agent to always successfully insert the peg into. Each episode lasts $50$ timesteps, which is only enough time to try one of the boxes. The agent's state consists of the robot's joint angles, joint velocities, and end-effector pose. The reward at each time-step is $1$ if the peg is successfully inserted into one of the boxes and $0$ otherwise.

Since this is a sparse-reward task, we first pre-train an agent on the dense rewards for $300$K environment steps with Soft Actor-Critic and save its replay buffer. We then initialize training of our method and of all comparisons with the restored replay buffer.
We train on $80$ different uncertainty sets, with $10$ sampled contexts from each set, for a total of $800$ different contexts.

\subsection{Experimental Results}
\label{app:experiments}

\begin{table*}
    \centering
    \setlength{\tabcolsep}{4pt}
    \begin{tabular}{cccc}
     & Task 1 & Task 2 & Task 3 \\
    System ID &
    \includegraphics[align=c,width=0.2\linewidth]{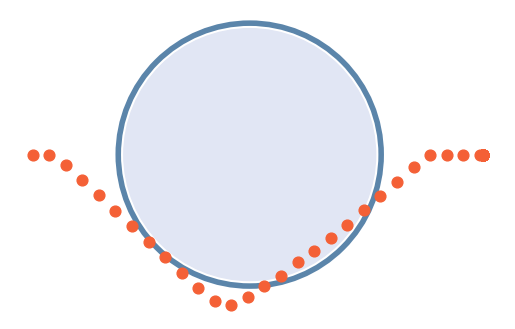} &
    \includegraphics[align=c,width=0.2\linewidth]{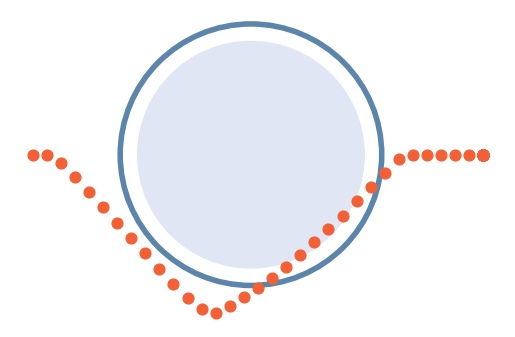} &
    \includegraphics[align=c,width=0.2\linewidth]{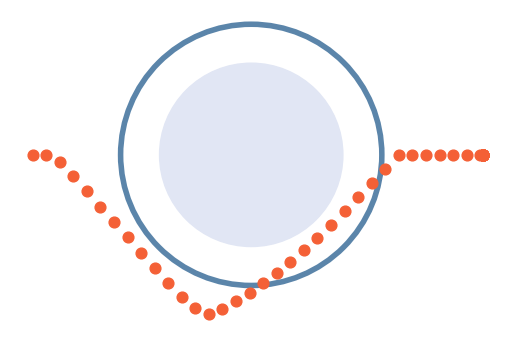} \\
    Set-EPOpt &
    \includegraphics[align=c,width=0.2\linewidth]{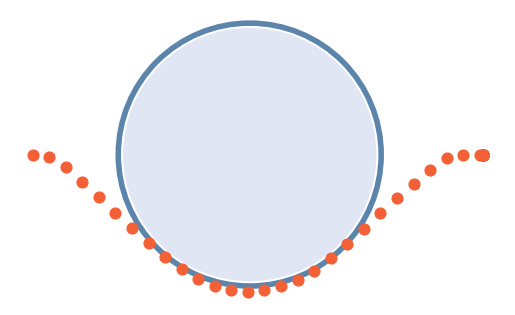} &
    \includegraphics[align=c,width=0.2\linewidth]{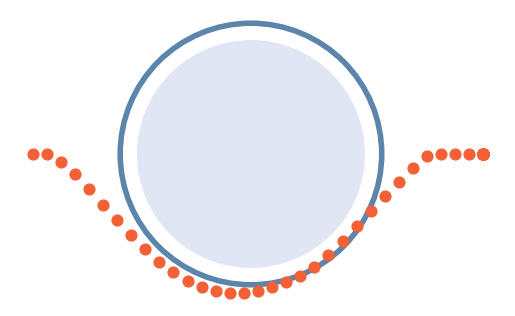} &
    \includegraphics[align=c,width=0.2\linewidth]{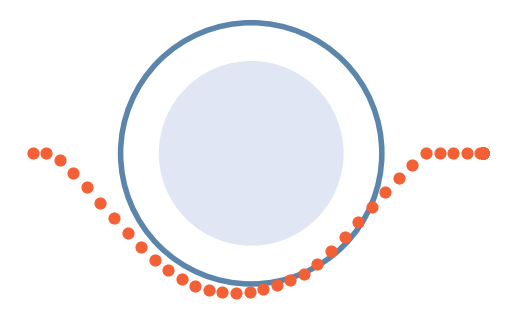} \\
    SIRSA (Ours) &
    \includegraphics[align=c,width=0.2\linewidth]{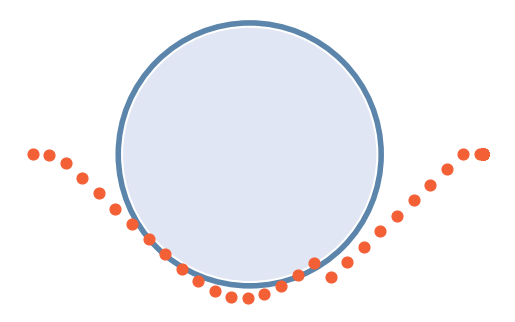} & \includegraphics[align=c,width=0.2\linewidth]{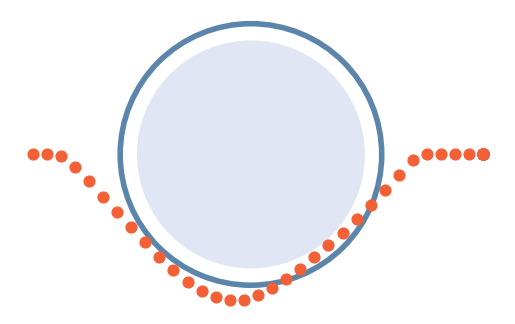} & \includegraphics[align=c,width=0.2\linewidth]{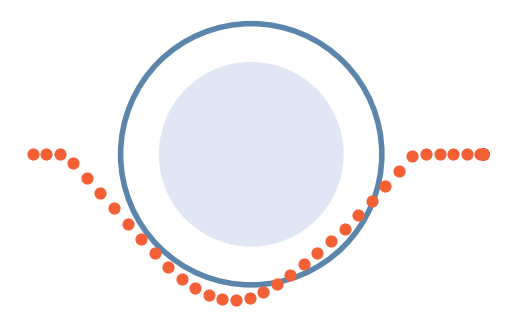}
    \end{tabular}
    \vspace{-0.2cm}
    \caption{\small Visualizations of the trajectories taken by policies learned with System ID, Set-EPOpt, and SIRSA in the Point Mass domain. The maximum obstacle size of this particular uncertainty set is demarcated by the unfilled blue circle, while the true obstacle in the environment is shaded in purple. The trajectory taken by the agent (from left to right) is in orange.}
    \label{table:point_mass}
    \vspace{-0.2cm}
\end{table*}

\subsubsection{Point Mass Visualizations}
\label{app:point_mass}
In Table~\ref{table:point_mass}, we visualize the trajectories of policies learned by System ID, Set-EPOpt, and SIRSA ($\alpha=0.25$) in the Point Mass domain. The unfilled blue circle marks the maximum obstacle size of the particular uncertainty set, while the true obstacle is shaded in lavender. The agent always starts to the left of the obstacle and must reach the right side. We mark the trajectory in orange. The trajectories taken by System ID (top row) tend to be overly optimistic and make contact with the obstacle, incurring penalty. Meanwhile, the trajectories taken by Set-EPOpt (middle row) are always along the maximum obstacle of the uncertainty set, which can be overly conservative, e.g., in the third task (third column), when the true obstacle is smaller. Finally, SIRSA (bottom row) strikes a balance between the two. Specifically, in the first task (first column), the agent initially makes contact with the obstacle but corrects its trajectory thereafter. In the subsequent two tasks, the agent is not too conservative but avoids the obstacle most of the time.

\subsubsection{Performance for Different CVaR $\alpha$'s}
\label{app:alpha}

In Tables~\ref{tbl:results_alpha_1} and~\ref{tbl:results_alpha_2}, we report the full results for the $\alpha$-dependent methods, EPOpt, Set-EPOpt, WCPG, Set-WCPG, and SIRSA (Ours), for $\alpha$ values in $\{ 0.25, 0.50, 0.75, 1.00 \}$.

\subsubsection{Sensitivity Analysis}
\label{app:sensitivity}

\begin{figure}
    \centering
    \includegraphics[width=0.48\linewidth]{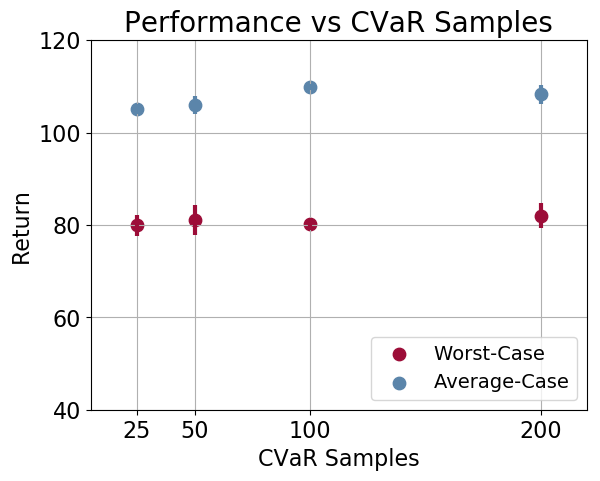}
    \includegraphics[width=0.48\linewidth]{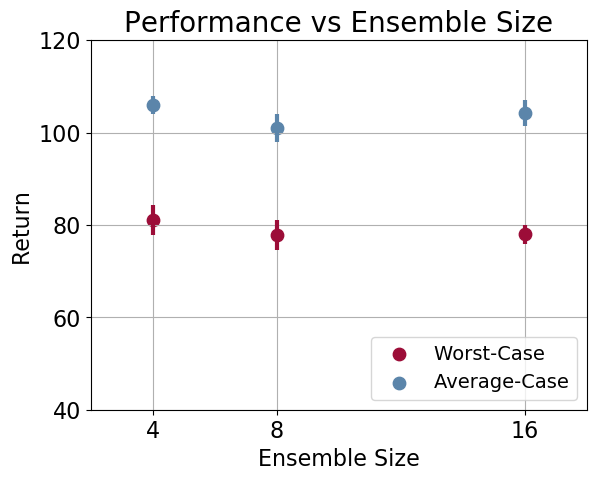}
    \vspace{-0.3cm}
    \caption{\small \textbf{Left:} Performance of SIRSA with different amounts of CVaR samples. \textbf{Right:} Performance of SIRSA with different ensemble sizes. Both plots depict the means and standard errors over $10$ random seeds.}
    \label{fig:sensitivity2}
\end{figure}

\textbf{Number of CVaR samples $N$.} Increasing the number of Monte-Carlo samples we use to approximate the CVaR objective can improve the estimate of the function. However, it is also more costly since it requires computing $\lfloor \alpha N \rfloor$ gradients. In Fig.~\ref{fig:sensitivity2} (left), we plot the average and worst-case performance of SIRSA for $N \in \{ 25, 50, 100, 200 \}$ in the Peg Insertion domain. From these results, we conclude that there is no significant benefit to increasing the CVaR samples beyond $50$.

\textbf{Number of ensemble models $B$.} Increasing the ensemble size can potentially lead to improved estimates of the posterior belief distribution, but requires training more model parameters. In Fig.~\ref{fig:sensitivity2} (right), we plot the average and worst-case performance for $B \in \{ 4,8,16 \}$ in the Peg Insertion domain. Overall, the performance of SIRSA is agnostic to the ensemble size in this domain.

\newpage
\begin{table}[]
    \centering
    \small
    \setlength{\tabcolsep}{3pt}
    \begin{tabular}{llccc}
        \toprule
             &        &   \\
        Task & Method & $\alpha$ & Min & Mean \\
        \midrule
        \multirow{20}{*}{Point mass} & \multirow{4}{*}{EPOpt} & $0.25$ & $34.3 \pm 0.6$  & $38.7 \pm 0.6$ \\
                           &  & $0.50$ & $35.2 \pm 0.6$ & $39.3 \pm 0.4$ \\
                           &  & $0.75$ & $36.1 \pm 0.6$ & $39.7 \pm 0.4$ \\
                           &  & $1.00$ & $36.3 \pm 0.6$ & $40.5 \pm 0.4$ \\
        \cmidrule{2-5}
        & \multirow{4}{*}{Set-EPOpt} & $0.25$ & $34.4 \pm 0.5$ & $38.5 \pm 0.4$ \\
                          &  & $0.50$ & $35.3 \pm 0.6$ & $39.2 \pm 0.5$ \\
                          &  & $0.75$ & $36.1 \pm 0.5$ & $40.1 \pm 0.3$ \\
                          &  & $1.00$ & $37.1 \pm 0.5$ & $40.7 \pm 0.4$ \\
        \cmidrule{2-5}
        & \multirow{4}{*}{WCPG} & $0.25$ & $34.8 \pm 0.6$ & $39.3 \pm 0.5$ \\
                          &  & $0.50$ & $34.7 \pm 0.6$ & $39.3 \pm 0.5$ \\
                          &  & $0.75$ & $34.6 \pm 0.6$ & $39.3 \pm 0.5$ \\
                          &  & $1.00$ & $34.4 \pm 0.6$ & $39.3 \pm 0.5$ \\
        \cmidrule{2-5}
        & \multirow{4}{*}{Set-WCPG} & $0.25$ & $34.7 \pm 0.7$ & $39.0 \pm 0.5$ \\
                         &  & $0.50$ & $34.6 \pm 0.7$ & $39.1 \pm 0.5$ \\
                         &  & $0.75$ & $34.6 \pm 0.7$ & $39.2 \pm 0.5$ \\
                         &  & $1.00$ & $34.6 \pm 0.7$ & $39.2 \pm 0.6$ \\
        \cmidrule{2-5}
        & \multirow{4}{*}{SIRSA (Ours)} & $0.25$ & $37.9 \pm 0.2$ & $41.8 \pm 0.1$ \\
                         &  & $0.50$ & $37.9 \pm 0.2$ & $41.7 \pm 0.1$ \\
                         &  & $0.75$ & $37.5 \pm 0.2$ & $41.9 \pm 0.1$ \\
                         &  & $1.00$ & $37.4 \pm 0.3$ & $41.3 \pm 0.3$ \\
                         
        \midrule
        \multirow{20}{*}{Minitaur} & \multirow{4}{*}{EPOpt} & $0.25$ & $131.8 \pm 10.7$ & $168.7 \pm 7.0$ \\
                           &  & $0.50$ & $160.1 \pm 8.4$ & $192.5 \pm 5.2$ \\
                           &  & $0.75$ & $171.3 \pm 7.4$ & $196.6 \pm 4.4$ \\
                           &  & $1.00$ & $172.2 \pm 7.3$ & $199.9 \pm 5.5$ \\
        \cmidrule{2-5}
        & \multirow{4}{*}{Set-EPOpt} & $0.25$ & $99.3 \pm 8.0$ & $141.2 \pm 6.2$ \\
                          &  & $0.50$ & $180.3 \pm 8.7$ & $211.5 \pm 3.6$ \\
                          &  & $0.75$ & $181.7 \pm 6.7$ & $213.9 \pm 2.4$ \\
                          &  & $1.00$ & $183.1 \pm 7.5$ & $216.7 \pm 3.7$ \\
        \cmidrule{2-5}
        & \multirow{4}{*}{WCPG} & $0.25$ & $163.0 \pm 17.4$ & $191.5 \pm 19.1$ \\
                          &  & $0.50$ & $163.5 \pm 17.6$ & $193.1 \pm 19.3$ \\
                          &  & $0.75$ & $163.3 \pm 18.1$ & $193.6 \pm 19.5$ \\
                          &  & $1.00$ & $165.5 \pm 17.7$ & $193.7 \pm 19.6$ \\
        \cmidrule{2-5}
        & \multirow{4}{*}{Set-WCPG} & $0.25$ & $173.1 \pm 9.2$ & $204.7 \pm 5.5$ \\
                         &  & $0.50$ & $173.4 \pm 9.1$ & $205.7 \pm 5.3$ \\
                         &  & $0.75$ & $173.1 \pm 9.3$ & $206.7 \pm 4.9$ \\
                         &  & $1.00$ & $174.5 \pm 10.0$ & $206.9 \pm 4.7$ \\
        \cmidrule{2-5}
        & \multirow{4}{*}{SIRSA (Ours)} & $0.25$ & $169.1 \pm 9.0$ & $205.4 \pm 3.8$ \\
                         &  & $0.50$ & $167.1 \pm 9.3$ & $199.6 \pm 3.5$ \\
                         &  & $0.75$ & $149.6 \pm 11.5$ & $192.7 \pm 5.0$ \\
                         &  & $1.00$ & $187.8 \pm 7.6$ &  $214.3 \pm 2.5$ \\ 
        \bottomrule
    \end{tabular}
    \caption{\small Evaluation on $20$ new uncertainty sets. Means and standard errors are computed over 10 random seeds for each method.}
    \label{tbl:results_alpha_1}
\end{table}

\begin{table}[]
    \centering
    \small
    \setlength{\tabcolsep}{3pt}
    \begin{tabular}{llccc}
        \toprule
             &        &   \\
        Task & Method & $\alpha$ & Min & Mean \\
        \midrule
        \multirow{20}{*}{Half-Cheetah} & \multirow{4}{*}{EPOpt} & $0.25$ & $1292 \pm 95$ & $1478 \pm 125$ \\
                           &  & $0.50$ & $1434 \pm 165$ & $1809 \pm 214$ \\
                           &  & $0.75$ & $1863 \pm 172$ & $2192 \pm 252$ \\
                           &  & $1.00$ & $2272 \pm 218$ & $2718 \pm 325$ \\
        \cmidrule{2-5}
        & \multirow{4}{*}{Set-EPOpt} & $0.25$ & $2148 \pm 42$ & $2454 \pm 47$ \\
                          &  & $0.50$ & $2845 \pm 29$ & $3274 \pm 93$ \\
                          &  & $0.75$ & $3246 \pm 206$ & $3703 \pm 171$ \\
                          &  & $1.00$ & $3811 \pm 224$ & $4474 \pm 232$ \\
        \cmidrule{2-5}
        & \multirow{4}{*}{WCPG} & $0.25$ & $3703 \pm 256$ & $4264 \pm 334$ \\
                          &  & $0.50$ & $3724 \pm 263$ & $4279 \pm 328$ \\
                          &  & $0.75$ & $3688 \pm 231$ & $4291 \pm 321$ \\
                          &  & $1.00$ & $3747 \pm 229$ & $4304 \pm 316$ \\
        \cmidrule{2-5}
        & \multirow{4}{*}{Set-WCPG} & $0.25$ & $3418 \pm 435$ & $3973 \pm 480$ \\
                         &  & $0.50$ & $3476 \pm 445$ & $3985 \pm 476$ \\
                         &  & $0.75$ & $3823 \pm 200$ & $4420 \pm 256$ \\
                         &  & $1.00$ & $3873 \pm 207$ & $4432 \pm 253$ \\
        \cmidrule{2-5}
        & \multirow{4}{*}{SIRSA (Ours)} & $0.25$ & $3742 \pm 209$ & $4288 \pm 228$ \\
                         &  & $0.50$ & $4126 \pm 72$ & $4806 \pm 75$ \\
                         &  & $0.75$ & $3583 \pm 298$ & $4103 \pm 280$ \\
                         &  & $1.00$ & $4146 \pm 112$ & $4872 \pm 73$ \\
                         
        \midrule
        \multirow{20}{*}{Peg Insertion} & \multirow{4}{*}{EPOpt} & $0.25$ & $0.0 \pm 0.0$ & $0.0 \pm 0.0$ \\
                           &  & $0.50$ & $11.4 \pm 7.6$ & $17.7 \pm 11.2$ \\
                           &  & $0.75$ & $43.2 \pm 5.4$ & $75.1 \pm 8.2$ \\
                           &  & $1.00$ & $41.3 \pm 5.3$ & $93.7 \pm 4.9$ \\
        \cmidrule{2-5}
        & \multirow{4}{*}{Set-EPOpt} & $0.25$ & $22.1 \pm 8.4$ & $33.3 \pm 11.2$ \\
                          &  & $0.50$ & $57.8 \pm 7.1$ & $78.0 \pm 9.5$ \\
                          &  & $0.75$ & $70.6 \pm 3.6$ & $96.9 \pm 2.7$ \\
                          &  & $1.00$ & $67.4 \pm 4.7$ & $77.8 \pm 8.3$ \\
        \cmidrule{2-5}
        & \multirow{4}{*}{WCPG} & $0.25$ & $20.2 \pm 5.4$ & $41.7 \pm 8.7$ \\
                          &  & $0.50$ & $28.8 \pm 6.3$ & $57.3 \pm 8.0$ \\
                          &  & $0.75$ & $33.8 \pm 7.2$ & $63.2 \pm 6.9$ \\
                          &  & $1.00$ & $31.8 \pm 5.9$ & $64.1 \pm 7.3$ \\
        \cmidrule{2-5}
        & \multirow{4}{*}{Set-WCPG} & $0.25$ & $56.6 \pm 4.5$ & $80.4 \pm 4.7$ \\
                         &  & $0.50$ & $64.0 \pm 4.2$ & $86.5 \pm 4.2$ \\
                         &  & $0.75$ & $66.4 \pm 4.0$ & $89.4 \pm 3.9$ \\
                         &  & $1.00$ & $68.3 \pm 4.6$ & $92.0 \pm 4.4$ \\
        \cmidrule{2-5}
        & \multirow{4}{*}{SIRSA (Ours)} & $0.25$ & $81.1 \pm 3.3$ & $106.0 \pm 1.9$ \\
                         &  & $0.50$ & $83.4 \pm 4.5$ & $109.5 \pm 2.6$ \\
                         &  & $0.75$ & $76.2 \pm 4.4$ & $102.1 \pm 3.4$ \\
                         &  & $1.00$ & $73.5 \pm 4.7$ & $103.1 \pm 2.3$ \\
        \bottomrule
    \end{tabular}
    \caption{\small Evaluation on $20$ new uncertainty sets. Means and standard errors are computed over 10 random seeds for each method.}
    \label{tbl:results_alpha_2}
\end{table}

\end{document}